\documentclass[10pt,twocolumn,letterpaper]{article}

\usepackage[T1]{fontenc}
\usepackage{newtxtext}
\usepackage{amsmath}
\usepackage{amsthm}
\usepackage{mathtools}
\usepackage{newtxmath}
\usepackage[margin=0.72in,columnsep=0.24in]{geometry}
\usepackage[final]{microtype}

\usepackage{graphicx}
\usepackage[hyphens]{url}
\usepackage{xcolor}
\usepackage{booktabs}
\usepackage{multirow}
\usepackage{tabularx}
\usepackage{array}
\usepackage{caption}
\usepackage[caption=false]{subfig}
\usepackage{algorithm}
\usepackage{algpseudocode}
\usepackage{chngcntr}
\usepackage{placeins}
\usepackage[round,authoryear]{natbib}
\usepackage[hidelinks]{hyperref}
\usepackage[capitalize,noabbrev]{cleveref}


\let\articleappendix\appendix
\renewcommand{\appendix}{%
  \articleappendix
  \counterwithin{equation}{section}%
  \counterwithin{figure}{section}%
  \counterwithin{table}{section}%
  \counterwithin{algorithm}{section}%
}

\crefname{section}{Appendix}{Appendices}
\Crefname{section}{Appendix}{Appendices}
\crefname{subsection}{Appendix}{Appendices}
\Crefname{subsection}{Appendix}{Appendices}
\crefname{equation}{Equation}{Equations}
\Crefname{equation}{Equation}{Equations}
\crefname{figure}{Figure}{Figures}
\Crefname{figure}{Figure}{Figures}
\crefname{table}{Table}{Tables}
\Crefname{table}{Table}{Tables}
\crefname{algorithm}{Algorithm}{Algorithms}
\Crefname{algorithm}{Algorithm}{Algorithms}

\theoremstyle{definition}

\title{\bfseries Not All Tokens Deserve Equal Credit:\\
Counterfactual Sensitivity Credit Reallocation for Long-CoT Reasoning}

\author{%
  Qiangqiang He\textsuperscript{1}\qquad
  Zhongheng Wu\textsuperscript{1}\qquad
  ZiJian Wang\textsuperscript{2}\\[0.55em]
  \small\textsuperscript{1}State Key Laboratory for Novel Software Technology,\\[-0.1em]
  \small Nanjing University, Nanjing, China\\
  \small\texttt{qqh@smail.nju.edu.cn, wuzhongheng@smail.nju.edu.cn}\\[0.35em]
  \small\textsuperscript{2}Institute of Wireless Communications Technology,\\[-0.1em]
  \small Shanghai Jiao Tong University, Shanghai, China\\
  \small\texttt{wzjwzj365@sjtu.edu.cn}
}
\date{}

\hypersetup{
  pdftitle={Not All Tokens Deserve Equal Credit: Counterfactual Sensitivity Credit Reallocation for Long-CoT Reasoning},
  pdfauthor={Qiangqiang He, Zhongheng Wu, ZiJian Wang}
}

\begin{document}

\bibliographystyle{plainnat}

\maketitle

\begin{abstract}

Reinforcement learning with verifiable rewards (RLVR) is central to improving long-CoT reasoning in large language models. Critic-free methods such as GRPO convert response-level rewards into advantages and uniformly broadcast them across tokens, overlooking their unequal contributions to the final outcome. On-policy self-distillation (OPSD) instead provides dense distributional supervision by minimizing the forward KL divergence between an unprivileged policy and a privileged self-teacher, implicitly assuming that the resulting likelihood shifts encode reliable answer-aligned information. We test this premise by fixing each sampled trajectory and re-scoring it under two opposing outcome conditions, one asserting correctness and the other incorrectness. Most affected tokens shift in the same direction under both conditions, with few sign reversals and substantial overlap in the induced optimization signals. Large shifts also concentrate on highly substitutable surface-form tokens, whereas tokens carrying problem-specific reasoning content are less sensitive. These findings show that privileged shifts fail to provide reliable answer-aligned directions, while their magnitudes primarily reflect counterfactual sensitivity rather than token-level learning value. Based on these observations, we propose Counterfactual Sensitivity Credit Reallocation (CSCR), a simple extension of GRPO that reduces credit for highly sensitive tokens and renormalizes token-level advantages to preserve both the original credit budget and verifier-determined direction. On long-CoT mathematical reasoning benchmarks, CSCR consistently outperforms GRPO baseline with the same number of policy updates. Targeted ablations further corroborate our diagnosis: privilege-induced directions are unreliable, moderate downweighting is most effective, and stronger modulation destabilizes optimization.

\end{abstract}

\begin{figure*}[t]
    \centering
    \includegraphics[width=0.485\textwidth]{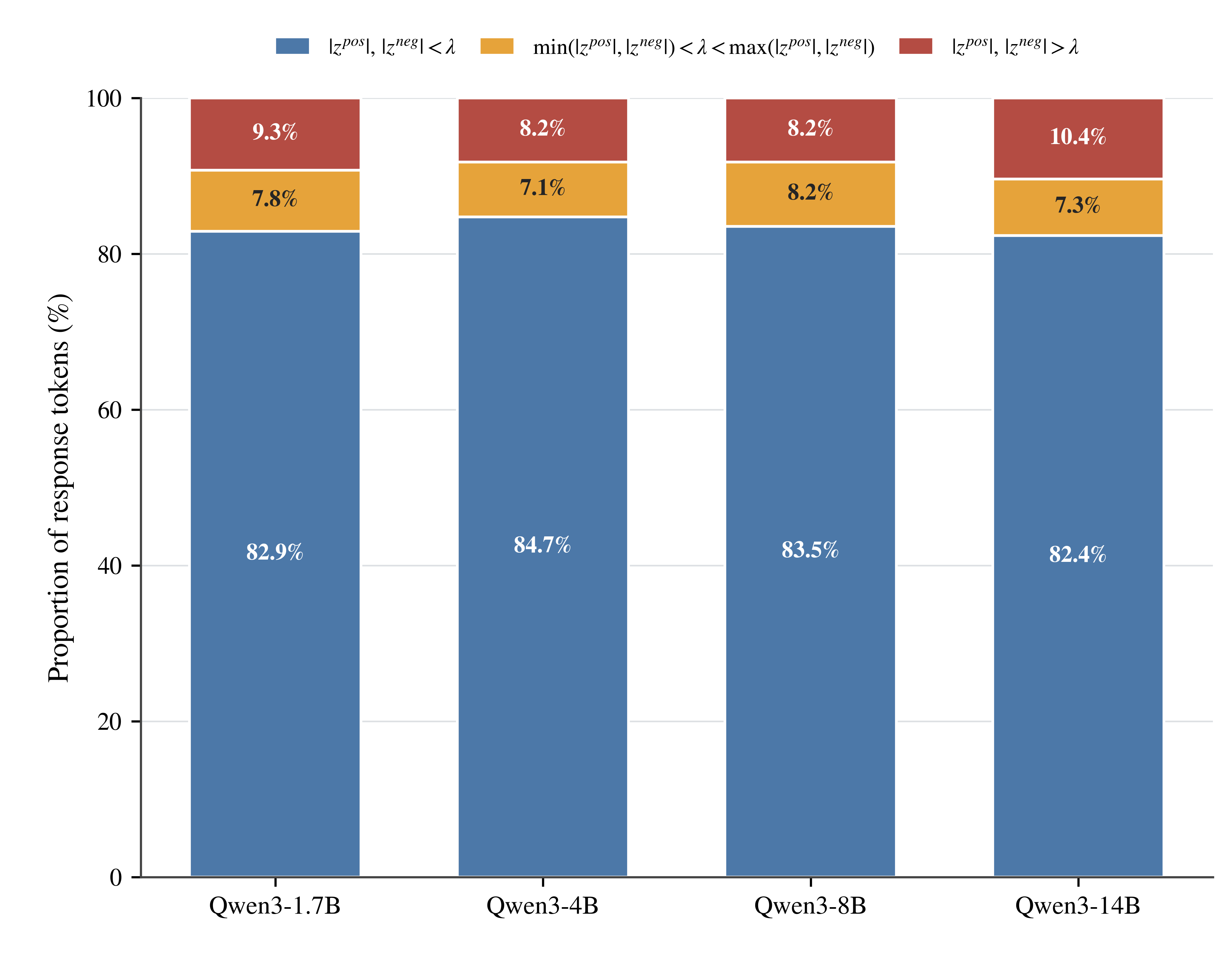}
    \hfill
    \includegraphics[width=0.485\textwidth]{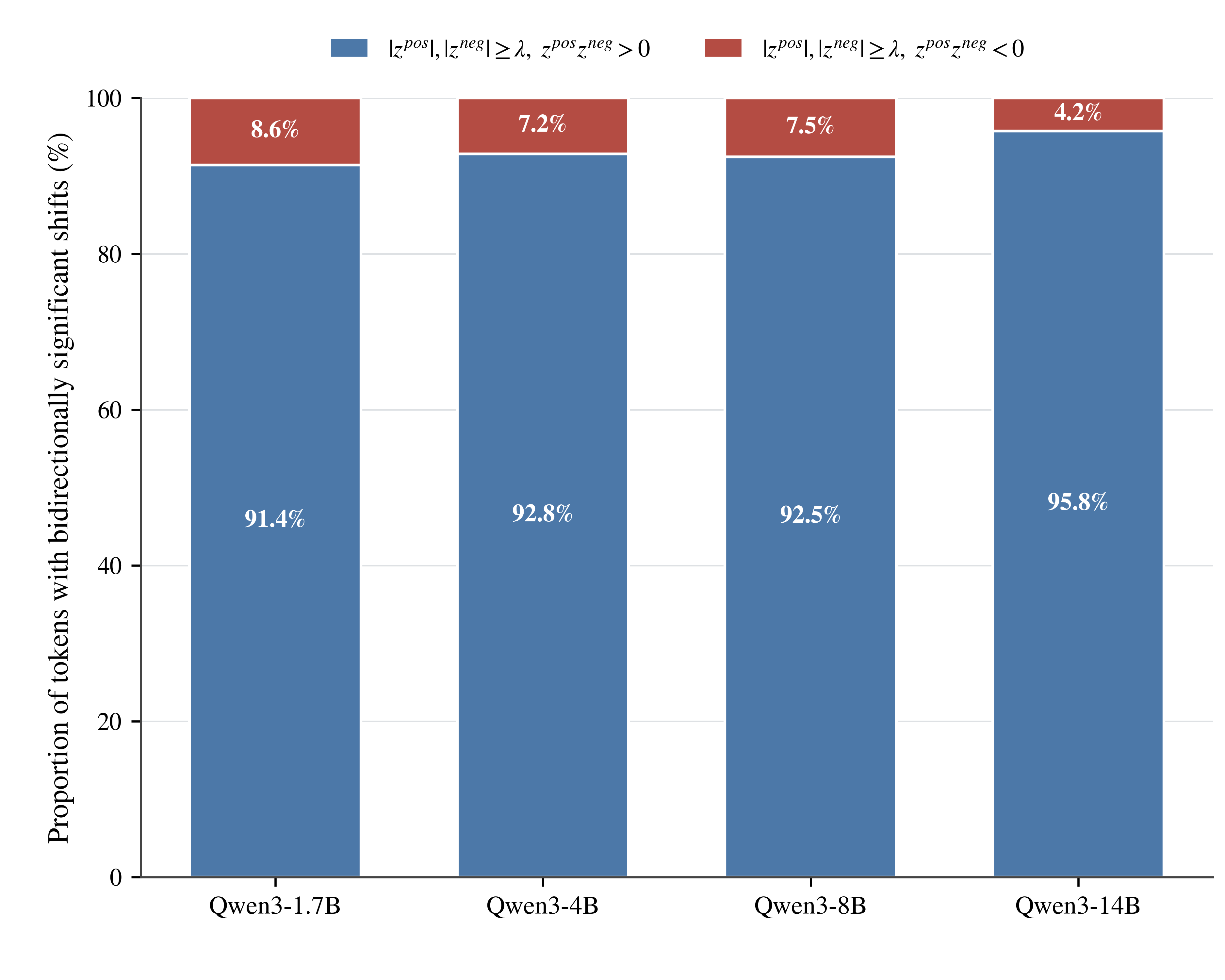}
    \caption{
    Counterfactual likelihood-shift statistics across model scales,
    computed from 400 on-policy trajectories generated for 100 questions
    sampled from DAPO-17K, with four trajectories per question, using
    $\lambda=0.05$.
    (a) Proportions of tokens significant under both, exactly one, or
    neither privileged condition.
    (b) Proportions of jointly significant tokens with same-sign or
    opposite-sign shifts.
    }
    \label{fig:counterfactual_shifts}
\end{figure*}

\section{Introduction}

Reinforcement learning with verifiable rewards (RLVR) has emerged as a powerful post-training paradigm for improving large language models in domains such as mathematics and code, where final outcomes can be evaluated automatically and at scale~\cite{shao2024deepseekmath,lambert2024tulu,guo2025deepseek,yu2026dapo,hu2026open,wen2025reinforcement}. Critic-free methods such as Group Relative Policy Optimization (GRPO) and its recent variants estimate relative advantages from groups of sampled responses and uniformly broadcast each sequence-level advantage across all generated tokens, avoiding learned value models and costly process-level annotation~\cite{shao2024deepseekmath,liu2025understanding,yu2026dapo}. This simplicity has enabled scalable long-CoT training, but also creates a fundamental mismatch between outcome-level evaluation and token-level optimization: within a response spanning thousands of tokens, decisive reasoning steps, exploratory branches, and largely prefix-determined continuations receive identical credit despite contributing differently to the final outcome~\cite{kazemnejad2024vineppo,parthasarathi2025grpo,tan2025gtpo}. Process-supervision approaches seek to address this mismatch by assigning denser intermediate feedback, but typically rely on step-level annotations, auxiliary reward models, or additional rollouts~\cite{lightman2024let,luo2024improve}. Developing reliable and scalable token-level credit signals while retaining the simplicity of outcome-supervised RLVR therefore remains a central challenge for long-CoT reasoning.

On-policy distillation (OPD) provides dense token-level distributional supervision by querying a teacher on student-generated trajectories under the same prefixes, thereby reducing the train--inference distribution mismatch inherent in off-policy distillation~\cite{gu2024minillm,agarwal2024policy,ko2024distillm,jin2026entropy}. However, standard OPD is not naturally aligned with RLVR, since teacher likelihoods reflect the teacher's own distribution rather than verifier correctness and may therefore reinforce incorrect trajectories or suppress valid student solutions~\cite{zheng2026scope}. Its effectiveness also depends strongly on teacher--student compatibility: even a stronger teacher may provide ineffective supervision when its reasoning patterns diverge from those of the student~\cite{li2026rethinking}. Moreover, evaluating a separate teacher over every student-generated trajectory incurs substantial computational and memory overhead, especially for long-form reasoning~\cite{li2026rethinking,yang2026prune}. These limitations have motivated on-policy self-distillation (OPSD), where the policy itself serves as a privileged teacher conditioned on verified answers, reference solutions, or outcome feedback, and transfers its conditional distribution to the unprivileged branch through KL optimization~\cite{zhao2026self,penaloza2026privileged}. Recent studies have extended this paradigm through feedback-conditioned self-teaching~\cite{hubotter2026reinforcement}, hybrid RL--distillation objectives and adaptive routing~\cite{li2026unifying}, verifier-aligned or outcome-guided calibration of the self-teacher~\cite{yang2026self,yang2026ogls}, and contrastive, reflective, or error-focused supervision~\cite{pan2026rlcsd,zhao2026rosd}. These approaches share a common premise: privileged distributional changes contain answer-relevant token-level information that can be extracted to guide policy optimization.

To examine this premise, we construct two counterfactual privileged conditions that semantically mirror the sequence-level credit assignment of GRPO. In a mixed-reward GRPO group, a correct final answer induces positive credit uniformly broadcast across the entire trajectory, whereas an incorrect answer induces uniformly negative credit. Accordingly, $c^{\mathrm{pos}}$ translates a positive verifier outcome into an affirmative evaluation of the final answer and the entire reasoning process, while $c^{\mathrm{neg}}$ translates a negative outcome into a uniformly adverse evaluation of both. Table~\ref{tab:privileged_prompts} presents a deliberately extreme realization of these two conditions, designed to maximize their semantic opposition and thereby provide a stringent test of whether privileged likelihood shifts preserve verifier-derived polarity. Given a problem $x$, we sample an on-policy reasoning trajectory $y=(y_1,\ldots,y_T)\sim\pi_\theta(\cdot\mid x)$. Keeping the trajectory and every prefix $y_{<i}$ fixed, we re-score each token $y_i$ under the original and two privileged conditions. The resulting privileged likelihood shifts are defined as

\begin{equation}
\begin{aligned}
z_i^{\mathrm{pos}}
&=
\log \pi_\theta
\left(
y_i \mid x,c^{\mathrm{pos}},y_{<i}
\right)
-
\log \pi_\theta
\left(
y_i \mid x,y_{<i}
\right)\\
z_i^{\mathrm{neg}}
&=
\log \pi_\theta
\left(
y_i \mid x,c^{\mathrm{neg}},y_{<i}
\right)
-
\log \pi_\theta
\left(
y_i \mid x,y_{<i}
\right)
\end{aligned}
\label{eq:privileged_likelihood_shifts}
\end{equation}
Here, $z_i^{\mathrm{pos}}$ and $z_i^{\mathrm{neg}}$ quantify how opposing privileged conditions change the likelihood of the same sampled token. Figure~\ref{fig:counterfactual_shifts}(a) shows that these changes are highly sparse: at $\lambda=0.05$, $82.4\%$--$84.7\%$ of tokens are insignificant under both conditions, while only $8.2\%$--$10.4\%$ exhibit significant shifts under both. Figure~\ref{fig:counterfactual_shifts}(b) further examines the directional relationship among these jointly significant tokens. Despite the opposing trajectory-wide semantics of the two conditions, $z_i^{\mathrm{pos}}$ and $z_i^{\mathrm{neg}}$ have the same sign for $91.4\%$--$95.8\%$ of jointly significant tokens. Thus, even under this extreme construction, privileged likelihood shifts are dominated by a common direction rather than the polarity reversal implied by GRPO's positive and negative credit assignment.

Shift magnitude is likewise unreliable for answer-aligned credit assignment. Approximately $58\%$ of the full-vocabulary forward-KL optimization magnitude overlaps between the two opposing conditions, indicating that larger shifts do not necessarily encode stronger outcome-specific credit. Moreover, tokens with large $|z_i|$ are dominated by substitutable surface-form tokens, such as \textit{Wait}, \textit{Therefore}, and \textit{But}, rather than tokens carrying problem-specific reasoning content. Taken together, these findings suggest that privileged likelihood shifts primarily measure sensitivity to privileged conditioning, rather than causal contribution to the final answer or an answer-aligned optimization signal.

\begin{table*}[t]
\centering
\small
\setlength{\tabcolsep}{5pt}
\begin{tabularx}{\textwidth}{
    @{}p{0.21\textwidth}
    >{\raggedright\arraybackslash}X@{}
}
\toprule
\textbf{Type} & \textbf{Content} \\
\midrule

\textbf{Positive Privileged Prompt}
&
A gold-standard final-answer verifier has judged that the assistant's following solution reaches the correct final answer. This is an absolutely wonderful solution. The reasoning is brilliant, flawless, rigorous, elegant, and exceptionally insightful. Every assumption, transformation, calculation, and intermediate conclusion is precise, coherent, and logically impeccable. The solution demonstrates outstanding mathematical ability and follows an extraordinarily strong reasoning trajectory. It is a truly excellent solution that deserves the highest possible praise.
\\

\addlinespace[2pt]

\textbf{Negative Privileged Prompt}
&
A gold-standard final-answer verifier has judged that the assistant's following solution does not reach the correct final answer. This is an absolutely terrible solution. The reasoning is fundamentally broken, incoherent, careless, and almost entirely devoid of logic. Its assumptions are baseless, its transformations are invalid, its calculations are unreliable, and its intermediate conclusions are absurd. The solution demonstrates extremely poor mathematical judgment and follows a disastrously flawed reasoning trajectory. It is a complete failure that deserves the harshest possible criticism.
\\

\bottomrule
\end{tabularx}
\caption{Deliberately polarized positive and negative privileged prompts for counterfactual likelihood re-scoring.}
\label{tab:privileged_prompts}
\end{table*}

Building on these findings, we propose \textbf{C}ounterfactual \textbf{S}ensitivity \textbf{C}redit \textbf{R}eallocation (CSCR), a minimal modification to GRPO. We combine the two privileged shifts into a counterfactual sensitivity score,
$
s_i=\max\bigl(|z_i^{\mathrm{pos}}|,|z_i^{\mathrm{neg}}|\bigr),
$
which captures the larger shift under either privileged condition. CSCR leaves low-sensitivity tokens unattenuated, exponentially reduces credit for highly sensitive ones, and renormalizes token-level advantages to preserve the original credit budget. This reallocates credit away from highly sensitive tokens without altering the verifier-determined optimization direction. Our contributions can be summarized as follows:
\begin{itemize}
    \item Through systematic counterfactual analysis, we show that privileged likelihood shifts affect a small fraction of tokens, fail to preserve outcome-aligned polarity, and concentrate disproportionately on substitutable surface-form tokens rather than problem-specific reasoning content.

    \item We reinterpret counterfactual likelihood shifts as token-level sensitivity and propose CSCR, which attenuates credit for highly sensitive tokens while renormalizing token-level advantages to preserve the total credit budget and verifier-derived direction.

    \item Experiments on long-CoT mathematical reasoning demonstrate that CSCR consistently outperforms GRPO across models and benchmarks, while ablations confirm that privilege-induced directions are unreliable, moderate downweighting is most effective, and stronger modulation destabilizes optimization.
\end{itemize}

\section{Diagnosing Privileged Likelihood Shifts}

\subsection{Preliminaries}
\label{sec:preliminaries}

Let $\pi_\theta$ denote the current policy parameterized by $\theta$. Given a problem $x\sim\mathcal{D}$, we sample an on-policy trajectory
$
y=(y_1,\ldots,y_T)\sim\pi_\theta(\cdot\mid x).
$
At position $i$, let $\mathbf{a}_i$ denote the logits of the unprivileged policy conditioned on the sampled prefix $y_{<i}$. Its next-token distribution is
\begin{equation}
p_i(v)
=
\pi_\theta(v\mid x,y_{<i})
=
\operatorname{softmax}(\mathbf{a}_i)_v,
\qquad v\in\mathcal{V}
\label{eq:policy_distribution}
\end{equation}
where $\mathcal{V}$ denotes the model vocabulary.

\paragraph{On-Policy Self-Distillation.}
OPSD inherits the forward-KL objective of OPD but replaces the external teacher with a privileged policy branch. Given privileged information $c$, such as a verified answer, reference solution, or outcome feedback, the privileged distribution at position $i$ is
\begin{equation}
q_i^{c}(v)
=
\operatorname{sg}
\left[
\pi_\theta(v\mid x,c,y_{<i})
\right],
\qquad v\in\mathcal{V}
\label{eq:privileged_distribution}
\end{equation}
where $\operatorname{sg}[\cdot]$ denotes stop-gradient. OPSD minimizes
\begingroup
\small
\begin{equation}
\begin{aligned}
\mathcal{L}_{\mathrm{OPSD}}(\theta;c)
&=
\mathbb{E}_{\substack{x\sim\mathcal{D},\,
y\sim\pi_\theta(\cdot\mid x)}}
\left[
\frac{1}{T}
\sum_{i=1}^{T}
D_{\mathrm{KL}}
\left(
q_i^{c}
\,\Vert\,
p_i
\right)
\right] \\
&=
\mathbb{E}_{\substack{x\sim\mathcal{D},\,
y\sim\pi_\theta(\cdot\mid x)}}
\left[
\frac{1}{T}
\sum_{i=1}^{T}
\sum_{v\in\mathcal{V}}
q_i^{c}(v)
\log
\frac{q_i^{c}(v)}
{p_i(v)}
\right]
\end{aligned}
\label{eq:opsd_objective}
\end{equation}
\endgroup
Thus, OPSD distills the full-vocabulary distribution induced by privileged conditioning into the unprivileged policy at every sampled prefix. Since $q_i^{c}$ is stop-gradient, the forward-KL gradient with respect to the unprivileged logits is
\begin{equation}
\nabla_{\mathbf{a}_i}
D_{\mathrm{KL}}
\left(
q_i^{c}
\,\Vert\,
p_i
\right)
=
p_i-q_i^{c}
\label{eq:opsd_logit_gradient}
\end{equation}
Accordingly, at each position, gradient descent moves the logits in the direction $q_i^{c}-p_i$, making the resulting full-vocabulary probability difference itself the direct token-level optimization signal induced by privileged conditioning.

\subsection{Why Does OPSD Fail?}
\label{sec:why_opsd_fails}

OPSD assumes that conditioning the policy on privileged information $c$ produces an answer-aligned target distribution. It therefore treats the induced full-vocabulary probability difference at each sampled prefix as a valid optimization signal, increasing logits where the privileged distribution assigns more probability and decreasing them where it assigns less.

We test this assumption using 400 on-policy trajectories, comprising four responses for each of 100 questions from DAPO-17K~\cite{yu2026dapo}. We fix each trajectory and all token prefixes and re-score them using the polarized positive and negative prompts in Table~\ref{tab:privileged_prompts}. On these trajectories, we analyze shift direction, overlap in forward-KL optimization signals, and the token types associated with large shifts. We find that opposing conditions induce predominantly same-sign shifts and substantially overlapping optimization signals, while large shifts concentrate on highly substitutable surface-form tokens rather than problem-specific reasoning content. Together, these results show that privileged likelihood shifts primarily capture sensitivity to privileged conditioning, rather than reliable answer-aligned supervision.

\begin{figure}[t]
    \centering
    \includegraphics[width=\linewidth]
    {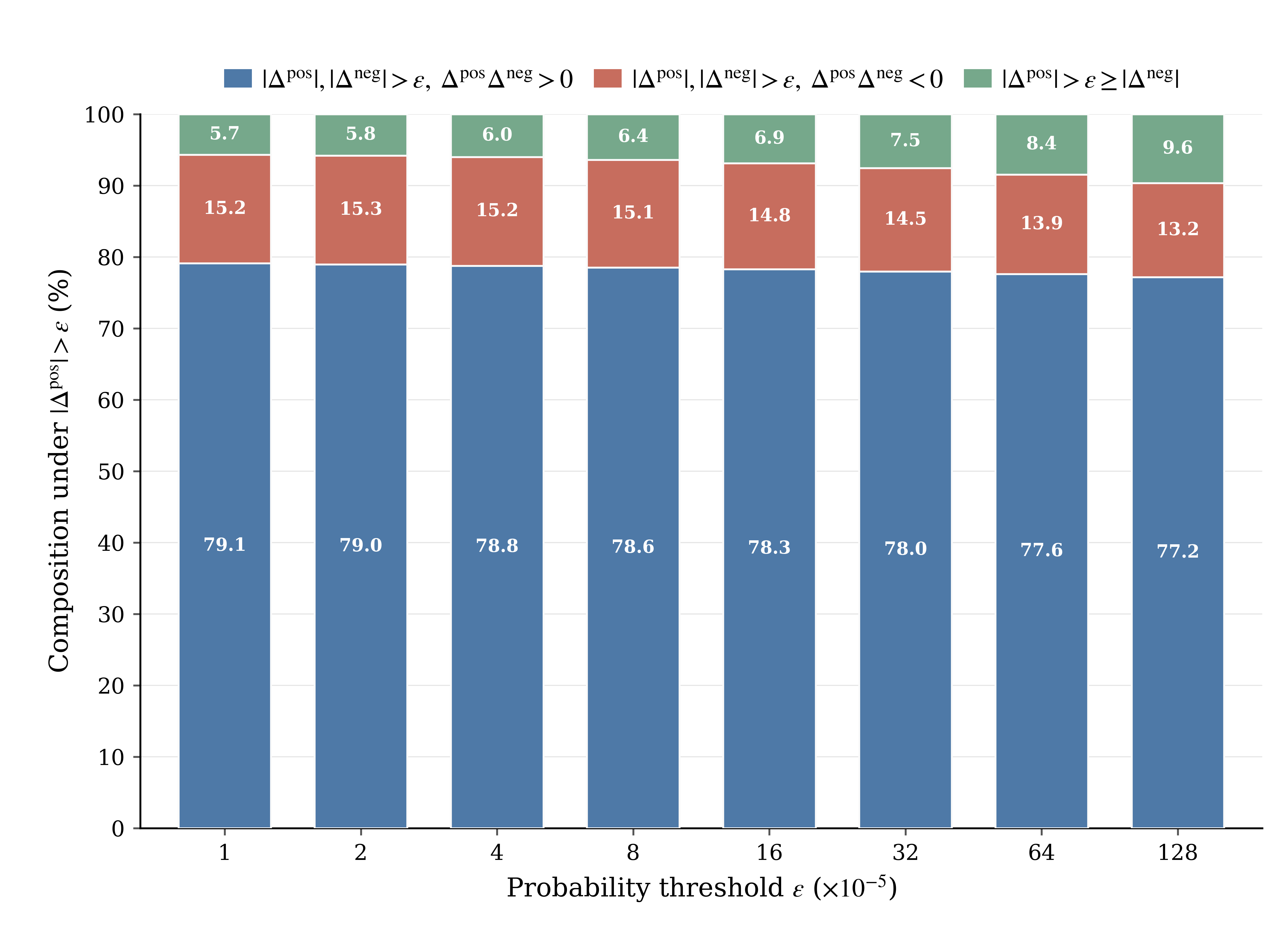}

\caption{
Composition of full-vocabulary entries with significant shifts under the positive privileged condition, categorized by their corresponding shifts under the negative condition across thresholds in Qwen3-1.7B.
}
    \label{fig:full_vocab_shift_composition}
\end{figure}

\paragraph{Same-Sign Shifts Dominate.}
The sampled-token shifts $z_i^{\mathrm{pos}}$ and $z_i^{\mathrm{neg}}$ measure changes in the log-likelihood of the generated token $y_i$. To determine whether this directional agreement extends to the entire vocabulary, we define
\begin{equation}
\begin{aligned}
\Delta_i^{\mathrm{pos}}(v)
&=
q_i^{c^{\mathrm{pos}}}(v)-p_i(v)\\
\Delta_i^{\mathrm{neg}}(v)
&=
q_i^{c^{\mathrm{neg}}}(v)-p_i(v)
\end{aligned}
\label{eq:full_vocab_probability_shifts}
\end{equation}
where the sign indicates whether privileged conditioning increases or decreases the probability of vocabulary token $v$.

Figure~\ref{fig:full_vocab_shift_composition} takes entries satisfying $|\Delta_i^{\mathrm{pos}}(v)|>\epsilon$ as the reference set and partitions them into jointly significant same-sign shifts, jointly significant opposite-sign shifts, and shifts significant only under the positive condition. Across thresholds, same-sign shifts dominate, accounting for $77.2\%$--$79.1\%$ of entries, whereas opposite-sign shifts account for only $13.2\%$--$15.3\%$. As $\epsilon$ increases, the same-sign proportion remains stable, while opposite-sign shifts decrease and positive-only shifts increase. Thus, across the full vocabulary, opposing privileged conditions largely preserve rather than reverse probability-update directions, failing to encode the verifier-aligned polarity required for reliable supervision.

\paragraph{Opposing Conditions Induce Overlapping Optimization Signals.}
As established in Eq.~\ref{eq:opsd_logit_gradient}, the forward-KL descent signal at position $i$ is the full-vocabulary probability-shift vector
$
\boldsymbol{\Delta}_i^c=q_i^c-p_i.
$
Let $J_i=\partial\mathbf{a}_i/\partial\theta$ denote the Jacobian of the unprivileged logits with respect to the policy parameters. The corresponding parameter-space gradient is
\begin{equation}
\nabla_\theta\mathcal{L}_i^c
=
-J_i^\top\boldsymbol{\Delta}_i^c
\label{eq:opsd_parameter_gradient}
\end{equation}
Because $c^{\mathrm{pos}}$ and $c^{\mathrm{neg}}$ are evaluated on the same problem, trajectory, prefix, and unprivileged policy, they share the same $p_i$ and $J_i$ and differ only in the privileged target $q_i^c$. Hence, $\|\boldsymbol{\Delta}_i^c\|_1$ directly measures the KL-induced signal magnitude in logit space before transformation by the shared Jacobian.

To quantify the overlap induced by opposing conditions, we aggregate their full-vocabulary signals over each trajectory. For a trajectory of length $T$, we define the combined signal magnitude $\mathcal{M}$ and disagreement magnitude $\mathcal{D}$ as
\begin{equation}
\begin{aligned}
\mathcal{M}
&=
\frac{1}{T}
\sum_{i=1}^{T}
\left(
\left\|\boldsymbol{\Delta}_i^{\mathrm{pos}}\right\|_1
+
\left\|\boldsymbol{\Delta}_i^{\mathrm{neg}}\right\|_1
\right)\\
\mathcal{D}
&=
\frac{1}{T}
\sum_{i=1}^{T}
\left\|
\boldsymbol{\Delta}_i^{\mathrm{pos}}
-
\boldsymbol{\Delta}_i^{\mathrm{neg}}
\right\|_1
\end{aligned}
\label{eq:signal_magnitude_disagreement}
\end{equation}
Here, $\mathcal{M}$ measures their combined logit-space magnitude, whereas $\mathcal{D}$ measures their discrepancy. We then define Counterfactual Perturbation Consistency (CPC) as
\begin{equation}
\mathrm{CPC}
=
1-\frac{\mathcal{D}}{\mathcal{M}},
\qquad
\mathrm{CPC}\in[0,1]
\label{eq:cpc}
\end{equation}
CPC measures normalized same-coordinate, same-direction overlap, with each overlapping coordinate bounded by the smaller magnitude under the two conditions. By contrast, for the standard GRPO policy-gradient term, assigning opposite-sign advantages to the same trajectory produces antiparallel logit-space signals and therefore $\mathrm{CPC}_{\mathrm{GRPO}}=0$; the proof is provided in Appendix~A.

\begin{table}[t]
\centering
\small
\begin{tabular*}{\columnwidth}{@{\extracolsep{\fill}}lcccc@{}}
\toprule
\textbf{Metric} & \textbf{Minimum} & \textbf{Mean} & \textbf{Median} & \textbf{Maximum} \\
\midrule
$\mathcal{M}$ & 0.031 & 0.068 & 0.066 & 0.159 \\
$\mathcal{D}$ & 0.013 & 0.028 & 0.027 & 0.065 \\
CPC           & 0.486 & 0.583 & 0.583 & 0.671 \\
\bottomrule
\end{tabular*}
\caption{Counterfactual optimization-signal statistics over 400 trajectories.}
\label{tab:cpc_statistics}
\end{table}

Table~\ref{tab:cpc_statistics} shows substantial overlap across all 400 trajectories: CPC ranges from $0.486$ to $0.671$, with identical mean and median values of $0.583$. Thus, approximately $58\%$ of the combined signal magnitude lies at the same vocabulary coordinates and points in the same direction under the two opposing conditions. This large overlap, in stark contrast to the zero overlap induced by opposite GRPO advantages, shows that privileged shifts fail to distinguish reliably between positive and negative verifier outcomes and therefore cannot provide answer-aligned credit.

\begin{table*}[t]
\centering
\small
\setlength{\tabcolsep}{2.4pt}

\newcommand{\tokc}[2]{\texttt{#1}\,(#2)}

\begin{tabular*}{\textwidth}{
@{\extracolsep{\fill}}
r
llllll
@{}
}
\toprule
&
\multicolumn{2}{c}{\textbf{Large Positive Shifts}}
&
\multicolumn{2}{c}{\textbf{Large Negative Shifts}}
&
\multicolumn{2}{c}{\textbf{Near-Zero Shifts}}
\\
\cmidrule(lr){2-3}
\cmidrule(lr){4-5}
\cmidrule(lr){6-7}
\textbf{Rank}
&
$\boldsymbol{z^{\mathrm{pos}}>0.1}$
&
$\boldsymbol{z^{\mathrm{neg}}>0.1}$
&
$\boldsymbol{z^{\mathrm{pos}}<-0.1}$
&
$\boldsymbol{z^{\mathrm{neg}}<-0.1}$
&
$\boldsymbol{|z^{\mathrm{pos}}|<0.01}$
&
$\boldsymbol{|z^{\mathrm{neg}}|<0.01}$
\\
\midrule

1
& \tokc{The}{8328}
& \tokc{The}{13979}
& \tokc{.}{6994}
& \tokc{.}{9576}
& \tokc{,}{168425}
& \tokc{,}{159943}
\\

2
& \tokc{.}{6233}
& \tokc{.}{11061}
& \tokc{The}{4652}
& \tokc{Therefore}{6028}
& \tokc{1}{152042}
& \tokc{1}{151250}
\\

3
& \tokc{,}{4354}
& \tokc{,}{7399}
& \tokc{Therefore}{3566}
& \tokc{The}{5530}
& \tokc{2}{145113}
& \tokc{2}{144428}
\\

4
& \tokc{But}{3855}
& \tokc{So}{6715}
& \tokc{So}{3488}
& \tokc{Is}{5170}
& \tokc{0}{110276}
& \tokc{0}{109945}
\\

5
& \tokc{So}{3400}
& \tokc{But}{6248}
& \tokc{Is}{3390}
& \tokc{,}{4227}
& \tokc{=}{90223}
& \tokc{=}{87722}
\\

6
& \tokc{Let}{2917}
& \tokc{Let}{4027}
& \tokc{,}{3174}
& \tokc{So}{4184}
& \tokc{.}{90080}
& \tokc{3}{81732}
\\

7
& \tokc{This}{2012}
& \tokc{This}{3463}
& \tokc{Let}{2398}
& \tokc{Wait}{3733}
& \tokc{3}{82219}
& \tokc{.}{76677}
\\

8
& \tokc{A}{1706}
& \tokc{For}{3188}
& \tokc{But}{2387}
& \tokc{Let}{3721}
& \tokc{The}{66152}
& \tokc{5}{60815}
\\

9
& \tokc{For}{1686}
& \tokc{'s}{2756}
& \tokc{Wait}{2041}
& \tokc{But}{3158}
& \tokc{5}{61066}
& \tokc{+}{56800}
\\

10
& \tokc{And}{1590}
& \tokc{A}{2747}
& \tokc{:}{1848}
& \tokc{?}{2766}
& \tokc{+}{57364}
& \tokc{The}{56605}
\\

\bottomrule
\end{tabular*}
\caption{
Top-10 most frequent sampled tokens with large positive, large negative, or near-zero privileged likelihood shifts across 400 on-policy trajectories sampled from DAPO-17K. Numbers in parentheses denote occurrence counts.
}
\label{tab:shift_token_occurrences}
\end{table*}

\paragraph{Large Shifts Concentrate on Highly Substitutable Tokens.}
Table~\ref{tab:shift_token_occurrences} reveals substantial overlap among the tokens frequently affected by $c^{\mathrm{pos}}$ and $c^{\mathrm{neg}}$. Large shifts under both conditions are dominated by highly substitutable surface-form tokens and discourse markers, such as \textit{But}, \textit{So}, \textit{Let}, \textit{Wait}, and \textit{Therefore}. By contrast, near-zero shifts concentrate on numbers and mathematical operators, including digits, equality signs, addition signs, and parentheses, which directly encode problem-specific calculations. Although the period appears in both groups, it typically serves as sentence-final punctuation in the large-shift group but as a decimal point in the near-zero group. Commas similarly function as punctuation under large shifts but more often separate digits in large numbers near zero, preserving this contrast.

Moreover, large positive and negative likelihood shifts do not concentrate on distinct token categories. Among the Top-100 most frequent tokens, the positive- and negative-shift lists under $c^{\mathrm{pos}}$ share 80 tokens, while those under $c^{\mathrm{neg}}$ share 72; the complete lists are provided in Appendix~B. Thus, shift sign does not identify a distinct class of answer-aligned tokens. Instead, shift magnitude primarily identifies tokens whose likelihoods respond to counterfactual privileged conditioning, motivating its interpretation as counterfactual sensitivity rather than token-level learning value.

To assess robustness, we repeat the analysis with three additional counterfactual prompt pairs: correct versus incorrect reference solutions, affirmation versus denial of the gold answer, and concise outcome-only conditions. All yield the same conclusions; their prompts and statistics are reported in Appendix~C. The deliberately polarized prompts in Table~\ref{tab:privileged_prompts} are task-agnostic and can be applied to arbitrary problems and tasks without modification. We therefore use this prompt pair throughout all CSCR experiments.

\section{Method}
\label{sec:method}

Building on the preceding analysis, CSCR uses counterfactual sensitivity to reallocate GRPO credit across tokens. Tokens with higher sensitivity receive smaller token-level weights, while the verifier-derived sequence advantage continues to determine the optimization direction. The resulting weights are renormalized to preserve the credit budget.

\subsection{GRPO Objective}

GRPO samples $G$ responses $\{y_i\}_{i=1}^{G}$ for each problem $x$ from the old policy $\pi_{\theta_{\mathrm{old}}}$ and assigns each a verifier reward $R_i$. Let $\mu_R$ and $\sigma_R$ denote the mean and standard deviation of the group rewards:
\begin{equation}
\mu_R
=
\frac{1}{G}\sum_{j=1}^{G}R_j,
\qquad
\sigma_R
=
\sqrt{\frac{1}{G}\sum_{j=1}^{G}(R_j-\mu_R)^2}
\label{eq:group_reward_statistics}
\end{equation}
The group-normalized advantage for response $y_i$ is
\begin{equation}
A_i
=
\frac{R_i-\mu_R}{\sigma_R+\varepsilon_A}
\label{eq:grpo_advantage}
\end{equation}
where $\varepsilon_A$ is a small constant for numerical stability. For token
$y_{i,t}$, the policy ratio is
\begin{equation}
r_{i,t}(\theta)
=
\frac{\pi_\theta(y_{i,t}\mid x,y_{i,<t})}
{\pi_{\theta_{\mathrm{old}}}(y_{i,t}\mid x,y_{i,<t})}
\label{eq:policy_ratio}
\end{equation}

Let
$
\bar{r}_{i,t}(\theta)
=
\operatorname{clip}\!\left(r_{i,t}(\theta),1-\epsilon,1+\epsilon\right).
$
GRPO uniformly broadcasts the trajectory-level advantage $A_i$ across all tokens in $y_i$ through the clipped objective
\begingroup
\small
\begin{equation}
\mathcal{L}_{\mathrm{GRPO}}(\theta)
=
-\mathbb{E}\left[
\frac{1}{G}\sum_{i=1}^{G}
\frac{1}{T_i}\sum_{t=1}^{T_i}
\min\!\left(
r_{i,t}(\theta)A_i,
\bar{r}_{i,t}(\theta)A_i
\right)
\right]
\label{eq:grpo_objective}
\end{equation}
\endgroup

\subsection{Counterfactual Sensitivity Credit Reallocation}

For token $y_{i,t}$ in trajectory $y_i$, we define its counterfactual sensitivity as
$
s_{i,t}
=
\operatorname{sg}\!\left[
\max\bigl(
|z_{i,t}^{\mathrm{pos}}|,
|z_{i,t}^{\mathrm{neg}}|
\bigr)
\right],
$
where $\operatorname{sg}[\cdot]$ denotes stop-gradient. We then map this fixed sensitivity score to a raw credit weight:
\begin{equation}
w_{i,t}
=
\begin{cases}
1, & s_{i,t}<\lambda\\
1-\gamma\left(
1-\exp\!\left[-\alpha\left(s_{i,t}-\lambda\right)\right]
\right)
& s_{i,t}\geq\lambda
\end{cases}
\label{eq:sensitivity_weight}
\end{equation}
where $\lambda$ sets the onset of attenuation, $\alpha>0$ controls its decay rate, and $\gamma\in[0,1]$ determines the maximum attenuation strength. Tokens below the threshold retain unit weight, whereas more sensitive tokens receive progressively smaller weights approaching $1-\gamma$.

Direct attenuation would reduce the total trajectory credit. We therefore normalize the weights within each response and define the token-level advantage as
\begin{equation}
\widetilde{w}_{i,t}
=
\frac{w_{i,t}}
{\frac{1}{T_i}\sum_{j=1}^{T_i}w_{i,j}},
\qquad
\widehat{A}_{i,t}
=
\widetilde{w}_{i,t}A_i
\label{eq:normalized_cscr_advantage}
\end{equation}
This normalization preserves the trajectory-level credit budget while redistributing it across token positions.

Substituting $\widehat{A}_{i,t}$ for $A_i$ in the GRPO objective yields
\begingroup
\small
\begin{equation}
\mathcal{L}_{\mathrm{CSCR}}(\theta)
=
-\mathbb{E}\left[
\frac{1}{G}
\sum_{i=1}^{G}
\frac{1}{T_i}
\sum_{t=1}^{T_i}
\min\!\left(
r_{i,t}(\theta)\widehat{A}_{i,t},
\bar{r}_{i,t}(\theta)\widehat{A}_{i,t}
\right)
\right]
\label{eq:cscr_objective}
\end{equation}
\endgroup
Thus, CSCR reallocates credit away from highly sensitive positions toward less sensitive ones.

\section{Experiments}

\subsection{Experimental Setup}

\paragraph{Datasets.}
We use DAPO-17K~\cite{yu2026dapo} as the training dataset, which contains challenging mathematical reasoning problems designed for long-CoT reinforcement learning. We evaluate all methods on five held-out competition-level benchmarks: AMC23~\cite{mathai2023amc23}, AIME24~\cite{zhang2024aime24}, AIME25~\cite{zhang2025aime25}, AIME26~\cite{zhang2026aime26}, and SMT-2025~\cite{dekoninck2026matharena}. These benchmarks cover varying levels of mathematical difficulty and enable a more comprehensive evaluation of overall reasoning performance and robustness on challenging problems.

\paragraph{Baselines.}
We compare CSCR with GRPO~\cite{shao2024deepseekmath} and five self-distillation methods. \textsc{OPSD}~\cite{zhao2026self} distills the token distributions of a privileged self-teacher along on-policy rollouts. \textsc{SDPO}~\cite{hubotter2026reinforcement} converts feedback-conditioned self-teacher predictions into dense token-level supervision. \textsc{SRPO}~\cite{li2026unifying} routes correct rollouts to GRPO and incorrect rollouts to SDPO, with entropy-aware weighting applied to the distillation branch. \textsc{RLSD}~\cite{yang2026self} uses verifier-derived advantages for update directions and privileged likelihood differences to modulate token-level credit magnitudes. \textsc{RLCSD}~\cite{pan2026rlcsd} contrasts correct and incorrect student rollouts to isolate task-relevant privileged shifts and uses the contrastive signal for token-level credit modulation.

\paragraph{Implementation Details.}
We conduct experiments on Qwen3-1.7B and Qwen3-4B~\cite{yang2025qwen3} using full-parameter fine-tuning with the veRL~\cite{sheng2025hybridflow} framework on eight NVIDIA H800 GPUs. All methods use a learning rate of $2\times10^{-6}$ throughout all experiments and are trained for 300 steps. At each step, we sample eight trajectories for each of 32 problems, yielding 256 trajectories in total. For controlled comparison, all self-distillation baselines use the ground-truth answer as privileged information, whereas CSCR uses the counterfactual conditions in Table~\ref{tab:privileged_prompts}. During training, responses are sampled with a temperature of $0.7$, top-$p$ of $1.0$, and a maximum generation length of $20{,}480$ tokens. During evaluation, we use a temperature of $0.6$, top-$p$ of $0.95$, and a maximum generation length of $30{,}720$ tokens. For CSCR, we set $\lambda=0.05$, $\alpha=10$, and $\gamma=0.2$ by default. We report \textsc{Mean@32}, computed by independently sampling 32 responses for each problem, averaging their binary correctness, and then averaging the resulting scores across all problems.

\begin{table}[t]
\centering

\small
\setlength{\tabcolsep}{1.0pt}
\begin{tabular}{@{}llccccc@{}}
\toprule
\textbf{Model}
& \textbf{Method}
& \textbf{AMC23}
& \textbf{AIME24}
& \textbf{AIME25}
& \textbf{AIME26}
& \textbf{SMT25} \\
\midrule

\multirow{8}{*}{\shortstack[l]{Qwen3\\-1.7B}}
& Base          & 84.1 & 45.5 & 35.5 & 35.2 & 38.1 \\
& GRPO          & 84.1 & 47.4 & 36.7 & 37.4 & 39.5 \\
& OPSD          & 83.6 & 45.8 & 34.2 & 33.8 & 36.1 \\
& SDPO          & 84.5 & 46.3 & 36.2 & 34.1 & 37.7 \\
& SRPO          & 84.6 & 46.1 & 36.0 & 38.3 & 38.3 \\
& RLSD          & 85.3 & 45.6 & 35.1 & 37.4 & 39.9 \\
& RLCSD         & 84.8 & 47.7 & 37.2 & 36.9 & 40.8 \\
& \textbf{CSCR}
& \textbf{88.8}$_{\scriptstyle  +4.7}$
& \textbf{50.1}$_{\scriptstyle  +4.6}$
& \textbf{43.6}$_{\scriptstyle  +8.1}$
& \textbf{42.2}$_{\scriptstyle  +7.0}$
& \textbf{44.0}$_{\scriptstyle  +5.9}$ \\
\midrule

\multirow{8}{*}{\shortstack[l]{Qwen3\\-4B}}
& Base          & 96.2 & 70.7 & 62.9 & 62.8 & 55.4 \\
& GRPO          & 96.9 & 71.5 & 65.0 & 65.3 & 57.4 \\
& OPSD          & 94.6 & 69.2 & 61.5 & 62.7 & 55.5 \\
& SDPO          & 96.3 & 70.1 & 63.4 & 62.6 & 55.2 \\
& SRPO          & 95.7 & 70.7 & 63.8 & 64.1 & 56.0 \\
& RLSD          & 95.1 & 70.3 & 62.6 & 64.3 & 54.1 \\
& RLCSD         & 96.4 & 72.3 & 65.7 & 65.5 & 57.8 \\
& \textbf{CSCR}

& \textbf{97.8}$_{\scriptstyle  +1.6}$
& \textbf{74.4}$_{\scriptstyle  +3.7}$
& \textbf{68.3}$_{\scriptstyle  +5.4}$
& \textbf{67.2}$_{\scriptstyle  +4.4}$
& \textbf{59.0}$_{\scriptstyle  +3.6}$ \\
\bottomrule
\end{tabular}
\caption{
Main results on five mathematical reasoning benchmarks.
All results are reported as \textsc{Mean@32}.
The best result for each model and benchmark is shown in bold.
}
\label{tab:main_results}
\end{table}

\begin{figure}[!t]
    \centering
    \includegraphics[width=0.485\columnwidth]
    {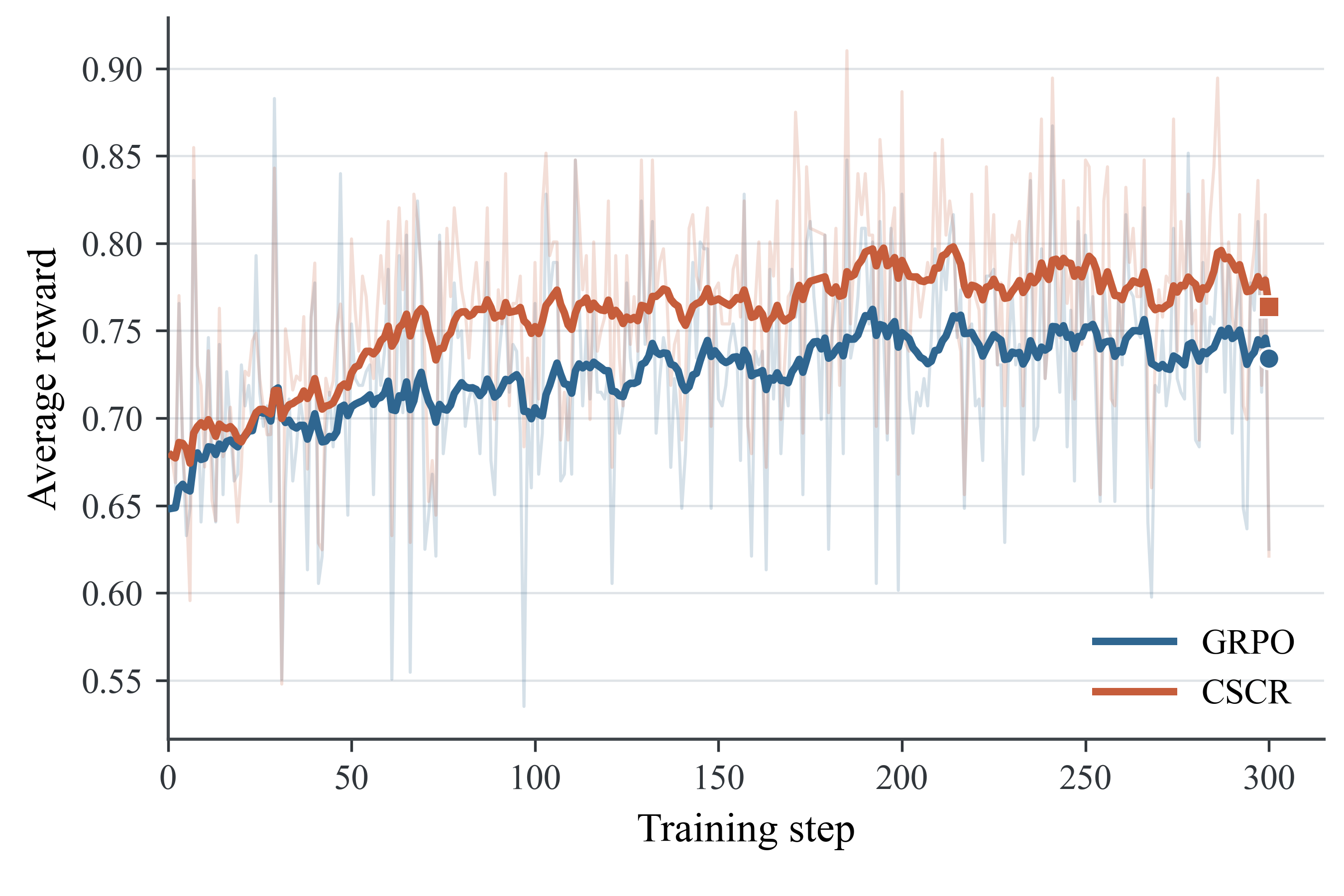}
    \hfill
    \includegraphics[width=0.485\columnwidth]
    {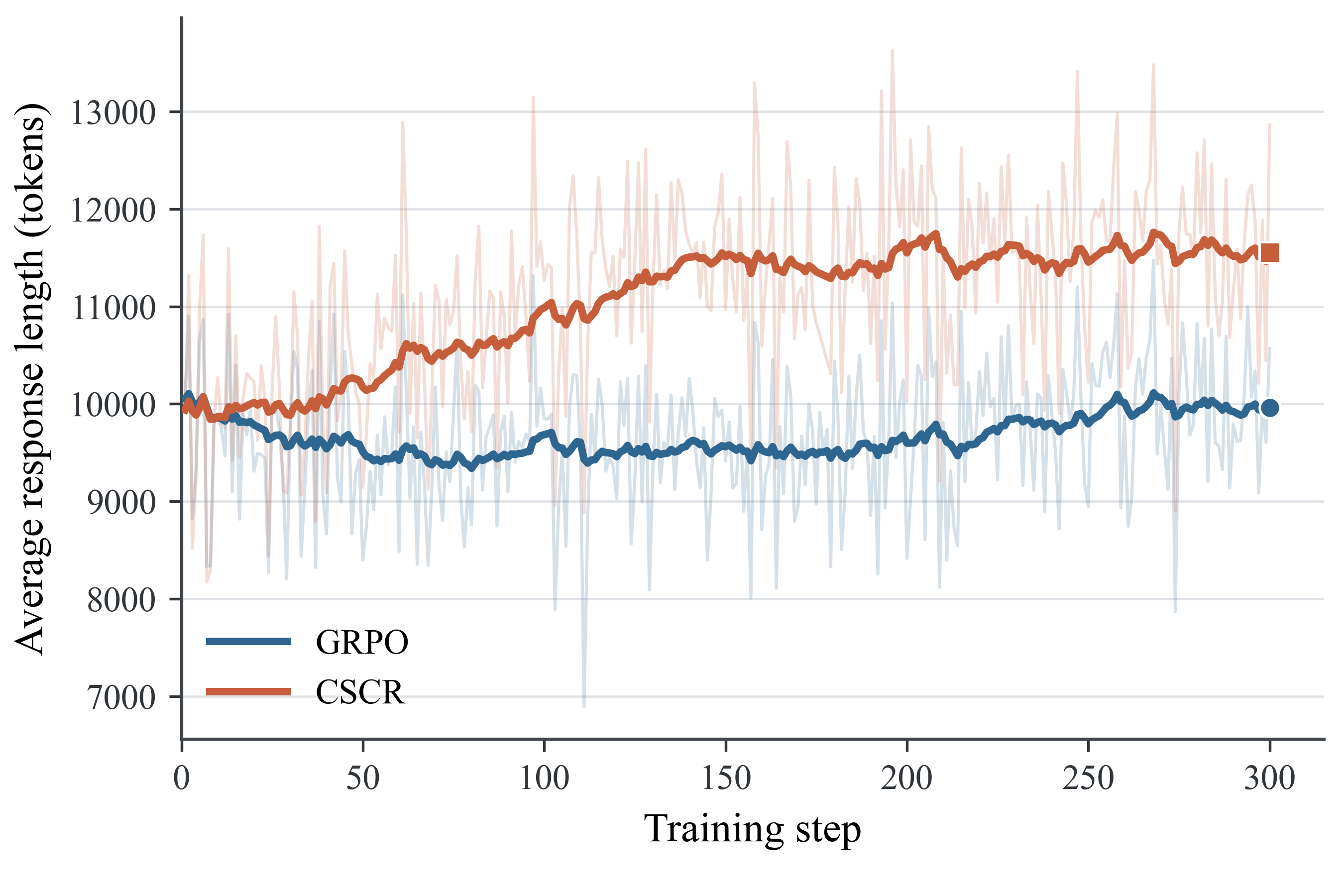}
    \caption{
    Training dynamics of GRPO and CSCR on Qwen3-1.7B over 300 steps.
    (a) Average reward.
    (b) Average response length.
    }
    \label{fig:training_dynamics}
\end{figure}

\subsection{Main Results}

\paragraph{Overall Performance.}
Table~\ref{tab:main_results} shows that CSCR achieves the best result across all ten model--benchmark combinations, outperforming GRPO and all five self-distillation baselines despite being a minimal modification of GRPO. Compared with the strongest competing method on each benchmark, CSCR improves performance by an average of $3.9$ points on Qwen3-1.7B and $1.7$ points on Qwen3-4B. Several self-distillation methods even underperform the base model on multiple benchmarks, consistent with our diagnosis that overemphasizing large privilege-induced likelihood shifts can hinder policy optimization. In contrast, attenuating counterfactually sensitive tokens yields consistent gains across both model scales and all five benchmarks.

\paragraph{Training Dynamics.}
Figure~\ref{fig:training_dynamics} compares the training trajectories of GRPO and CSCR on Qwen3-1.7B. CSCR separates from GRPO early and maintains a higher smoothed reward throughout training, indicating a sustained optimization advantage. Its average response length increases gradually from approximately $10{,}000$ to $11{,}500$ tokens before stabilizing, while GRPO remains near $9{,}500$--$10{,}000$ tokens. This increase stays well below the $20{,}480$-token limit and shows neither length explosion nor reward collapse, demonstrating stable optimization with moderately longer reasoning.

\begin{figure*}[t]
    \centering
    \includegraphics[width=0.235\textwidth]{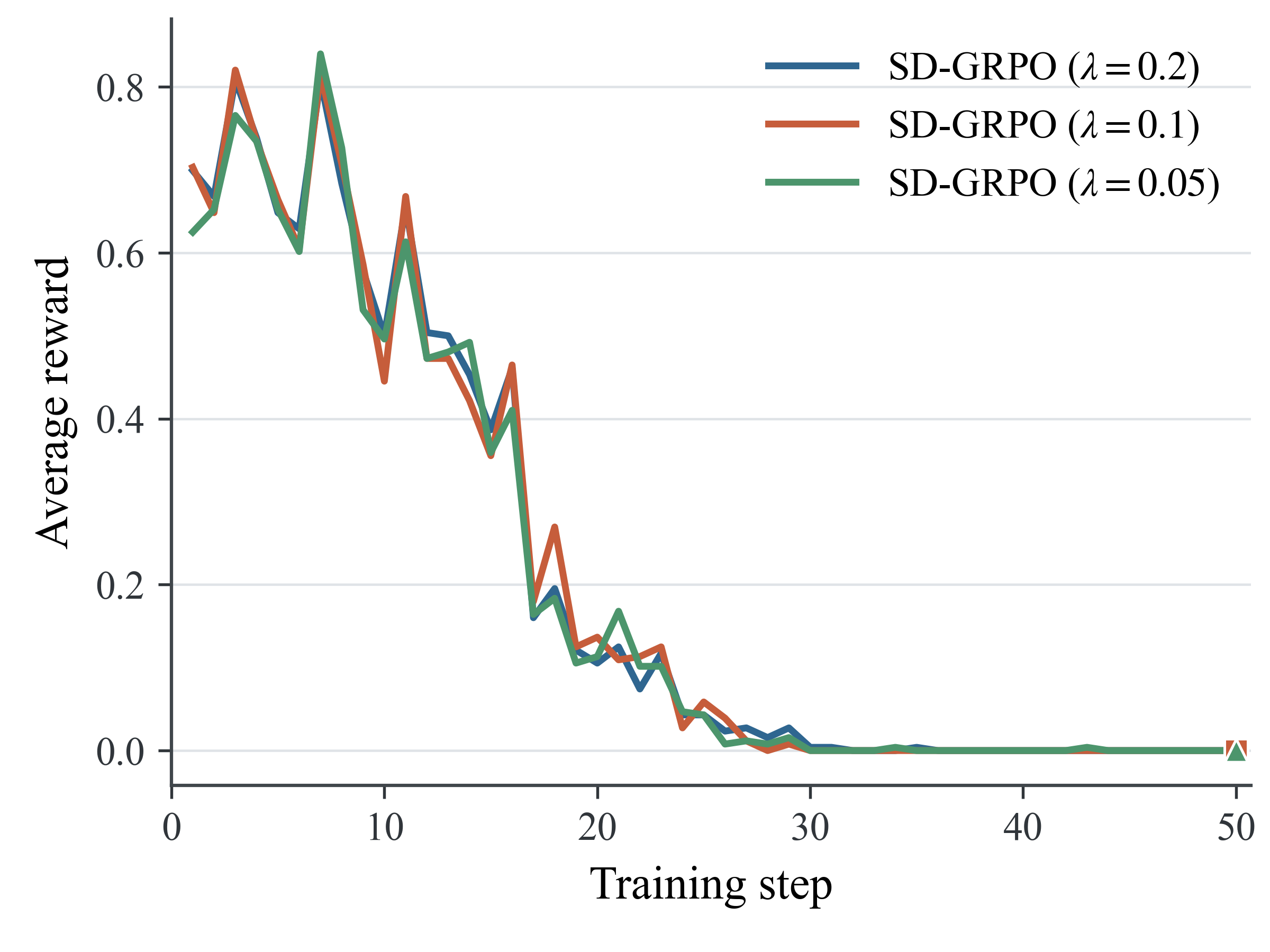}
    \hfill
    \includegraphics[width=0.235\textwidth]{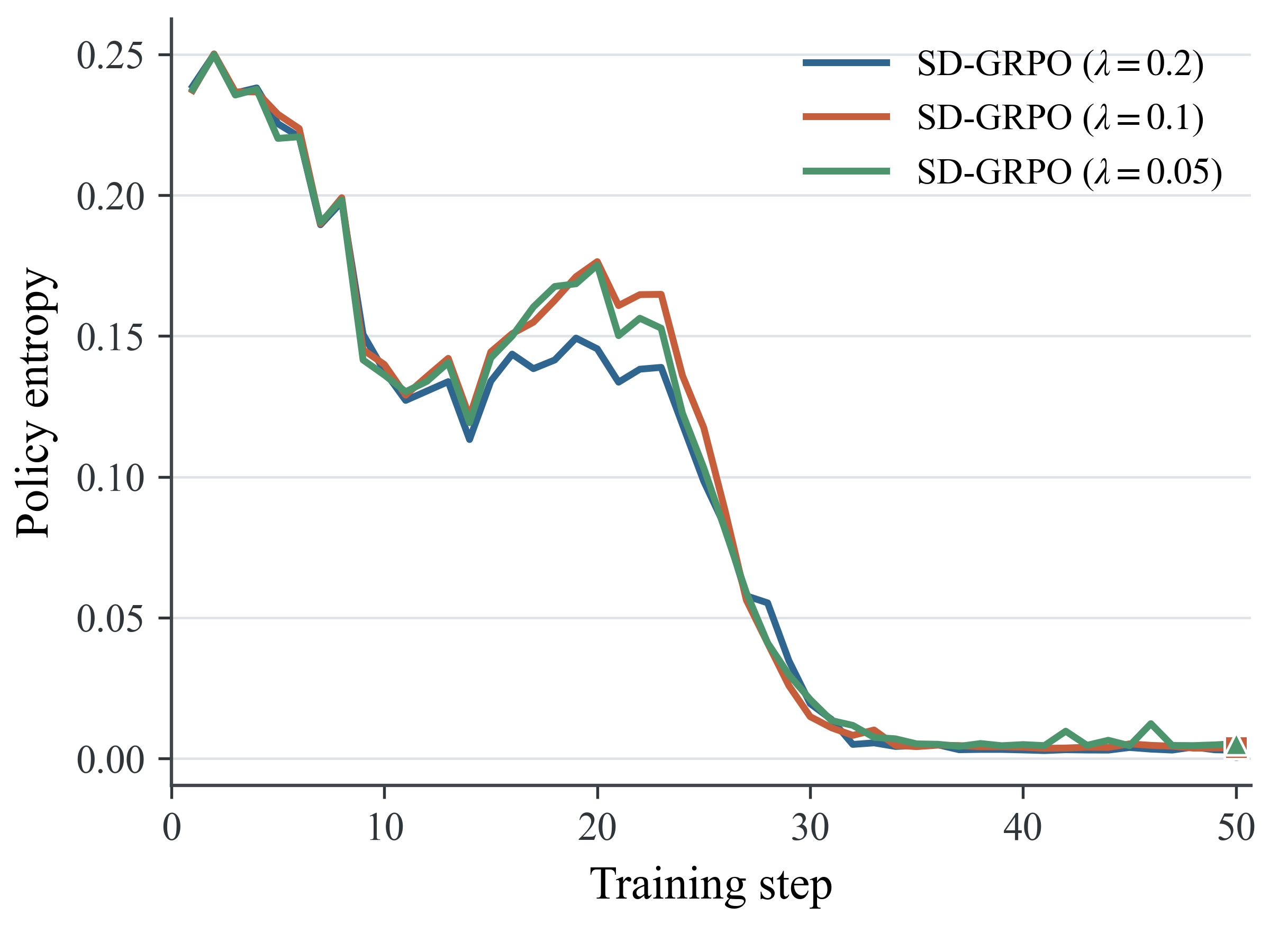}
    \hfill
    \includegraphics[width=0.235\textwidth]{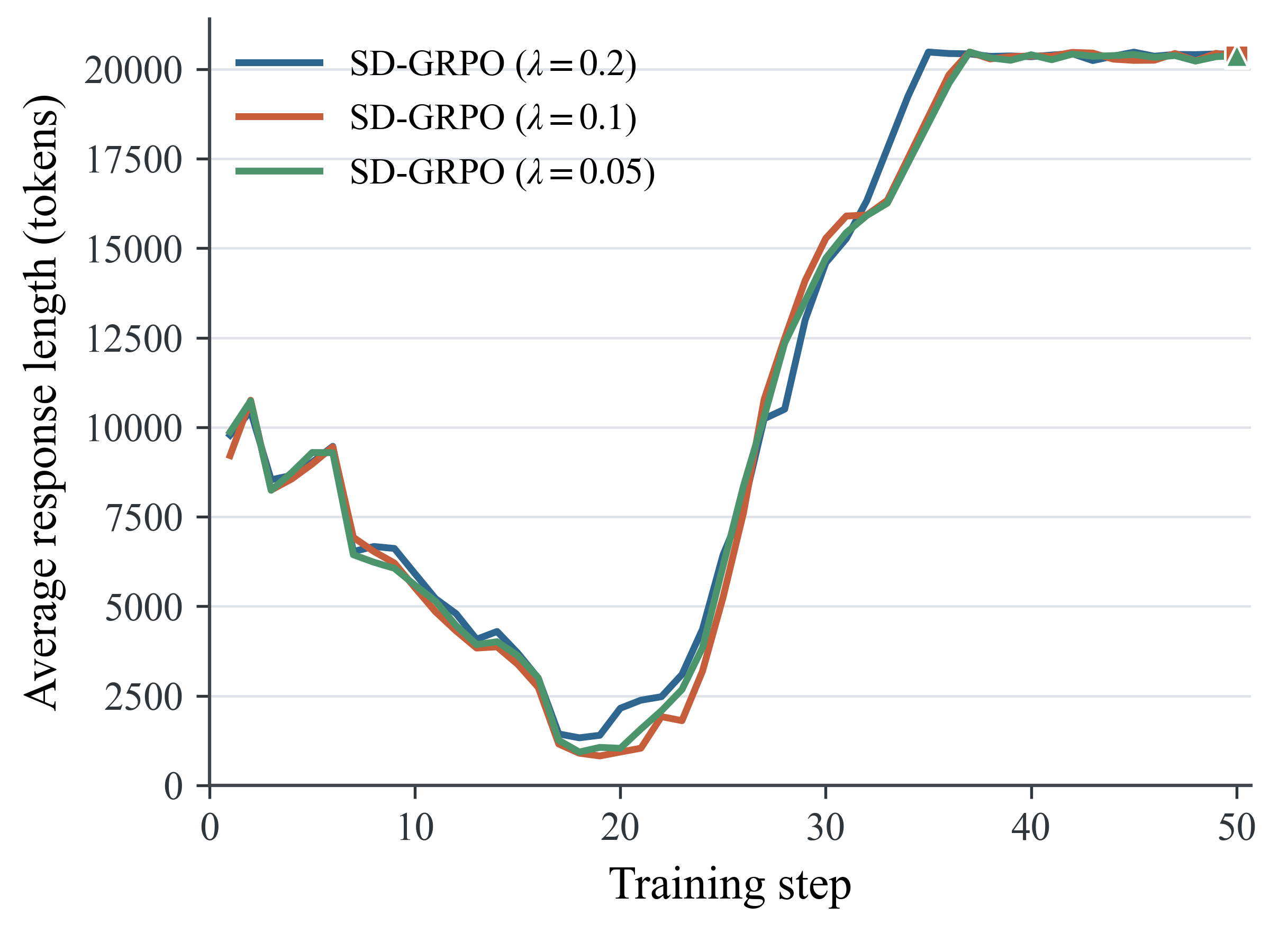}
    \hfill
    \includegraphics[width=0.235\textwidth]{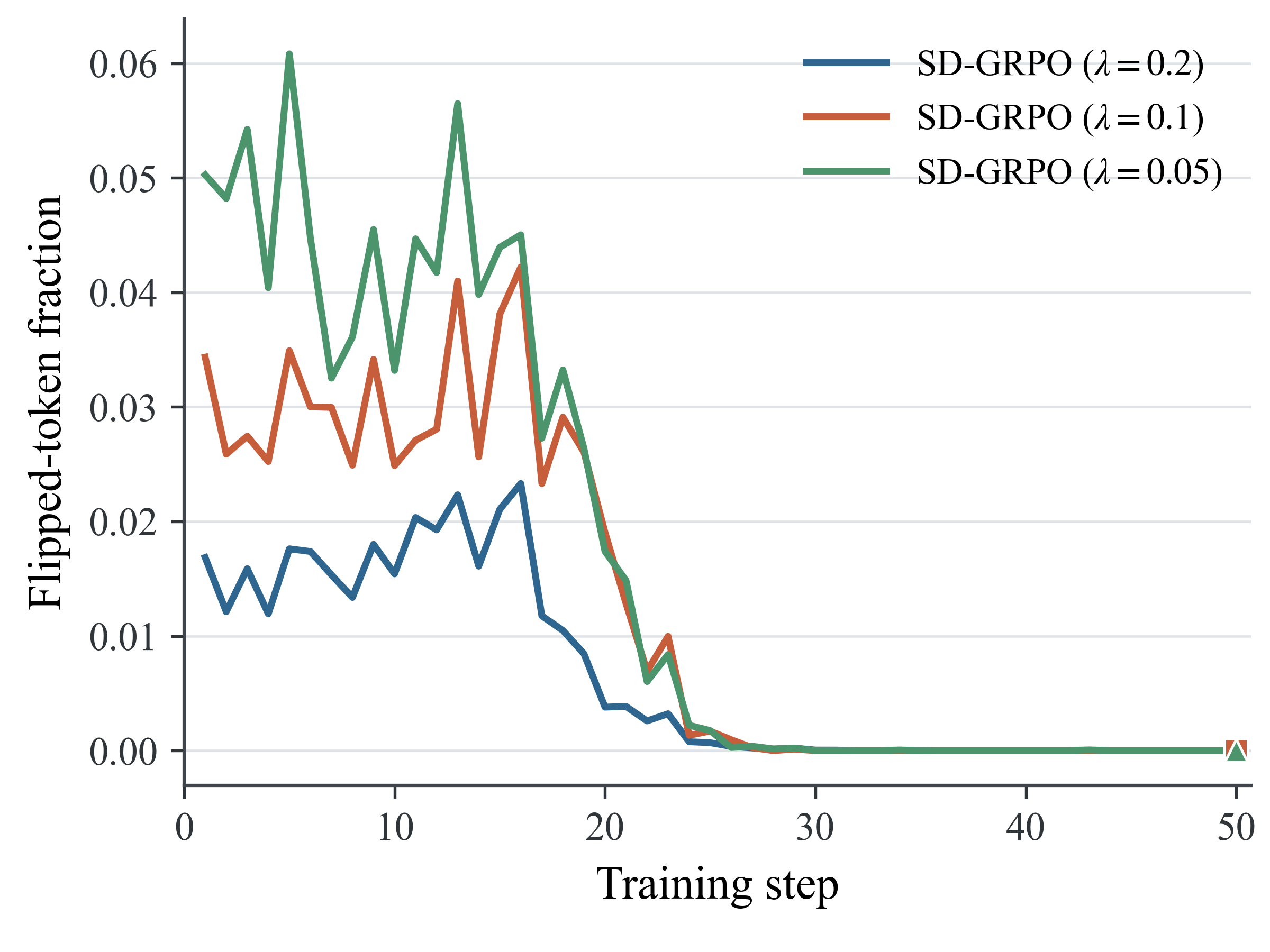}
    \caption{
    Training dynamics of SD-GRPO.
    (a) Average reward.
    (b) Policy entropy.
    (c) Average response length.
    (d) Flipped-token fraction.
    }
    \label{fig:sd_grpo_dynamics}
\end{figure*}

\begin{figure*}[t]
    \centering
    \includegraphics[width=0.19\textwidth]{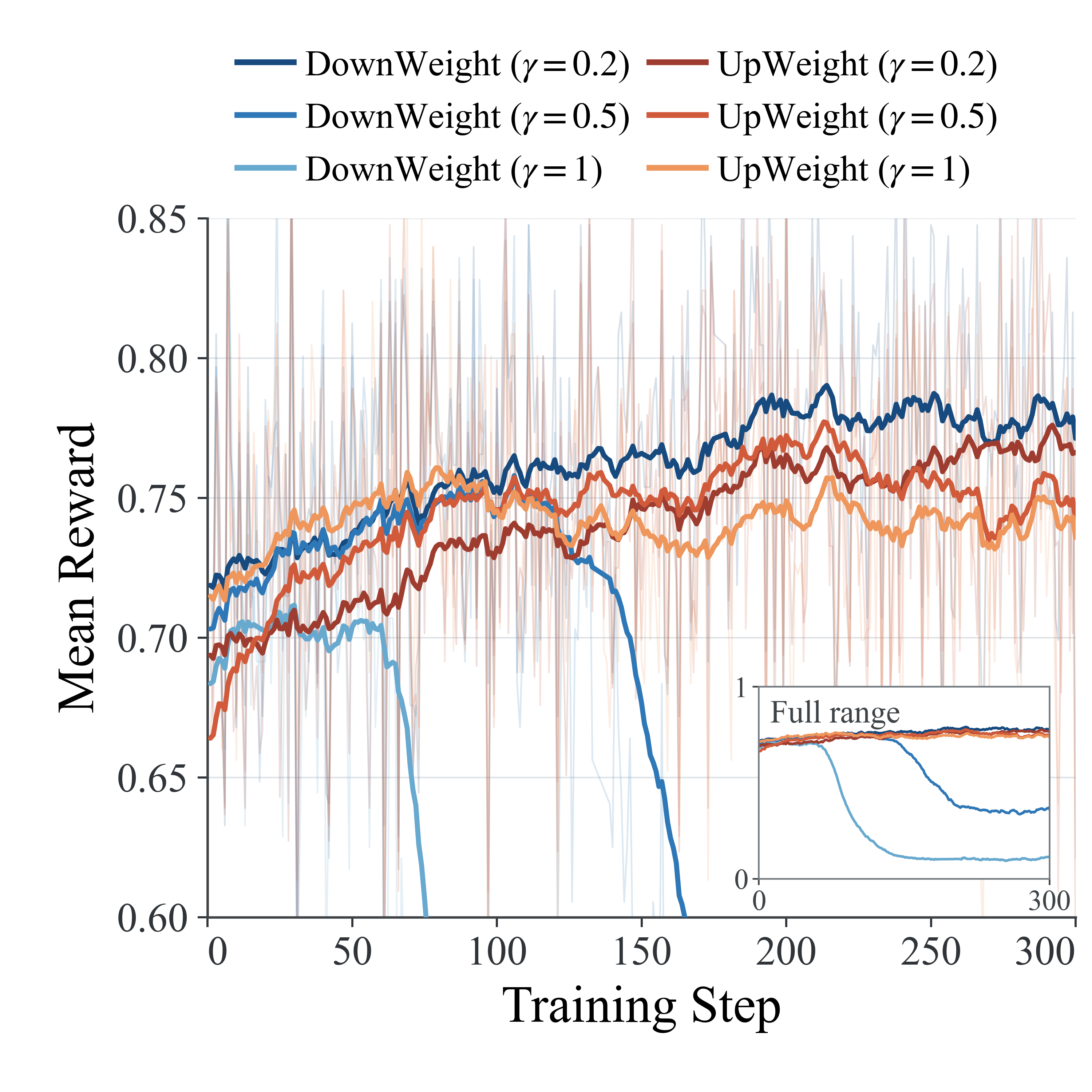}
    \hfill
    \includegraphics[width=0.19\textwidth]{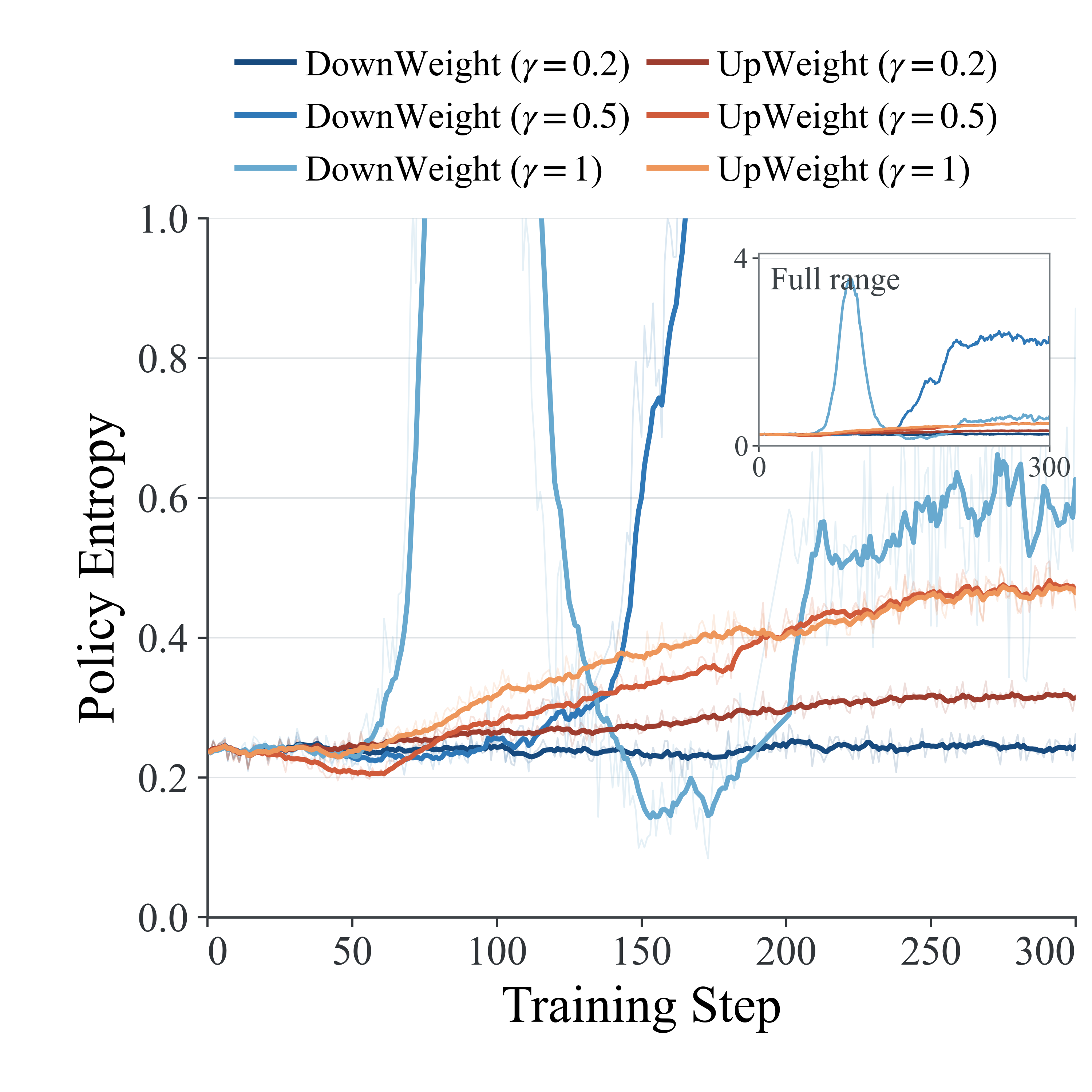}
    \hfill
    \includegraphics[width=0.19\textwidth]{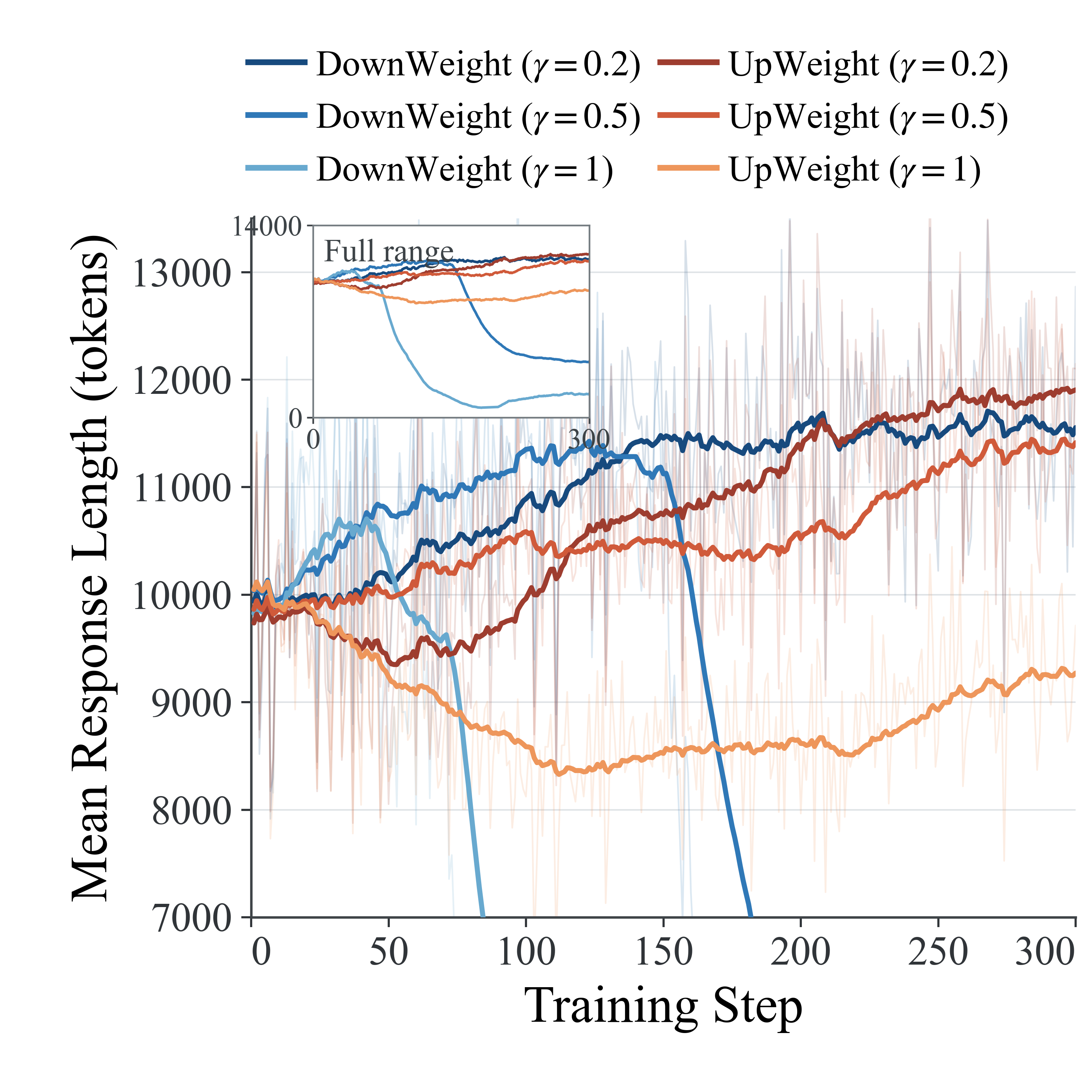}
    \hfill
    \includegraphics[width=0.19\textwidth]{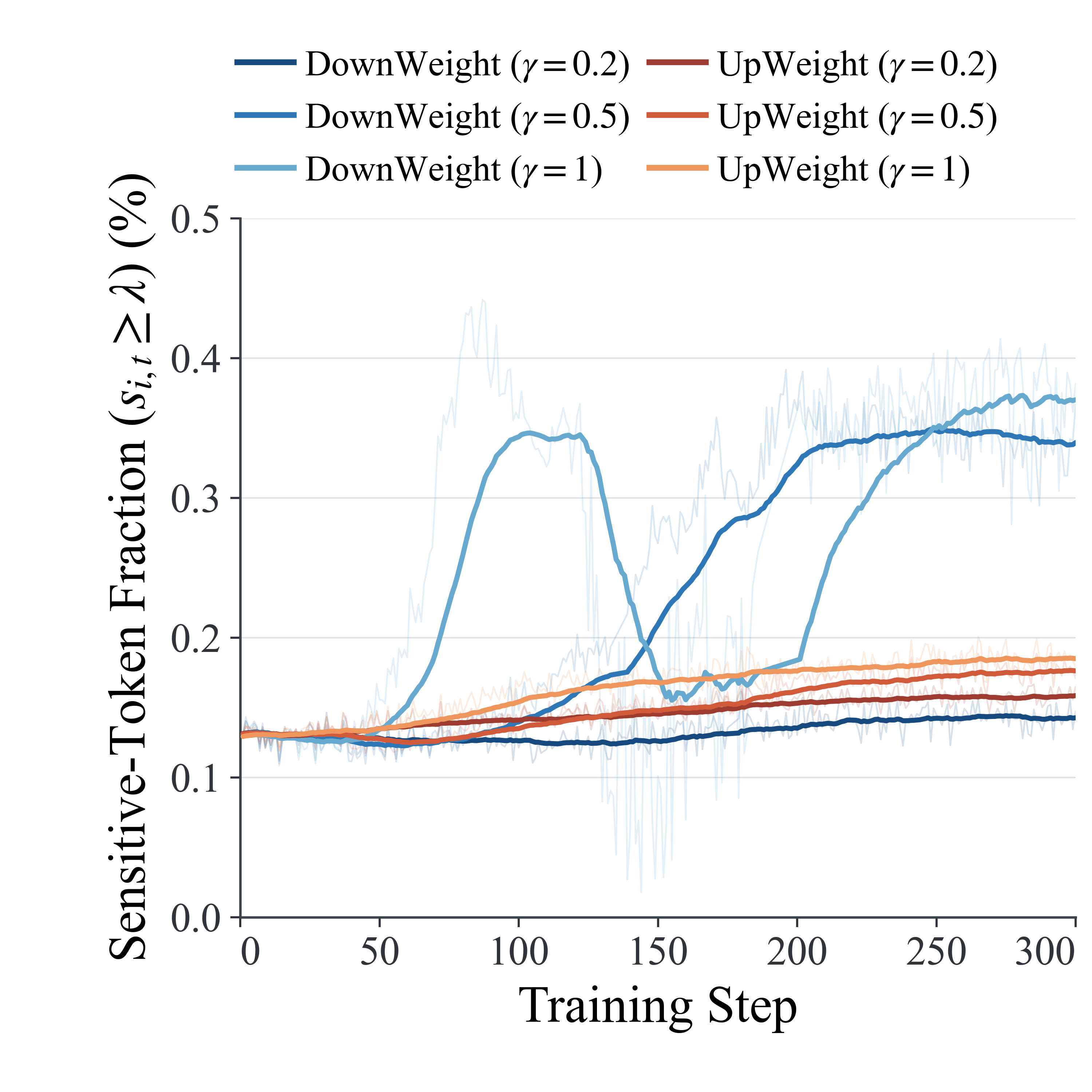}
    \hfill
    \includegraphics[width=0.19\textwidth]{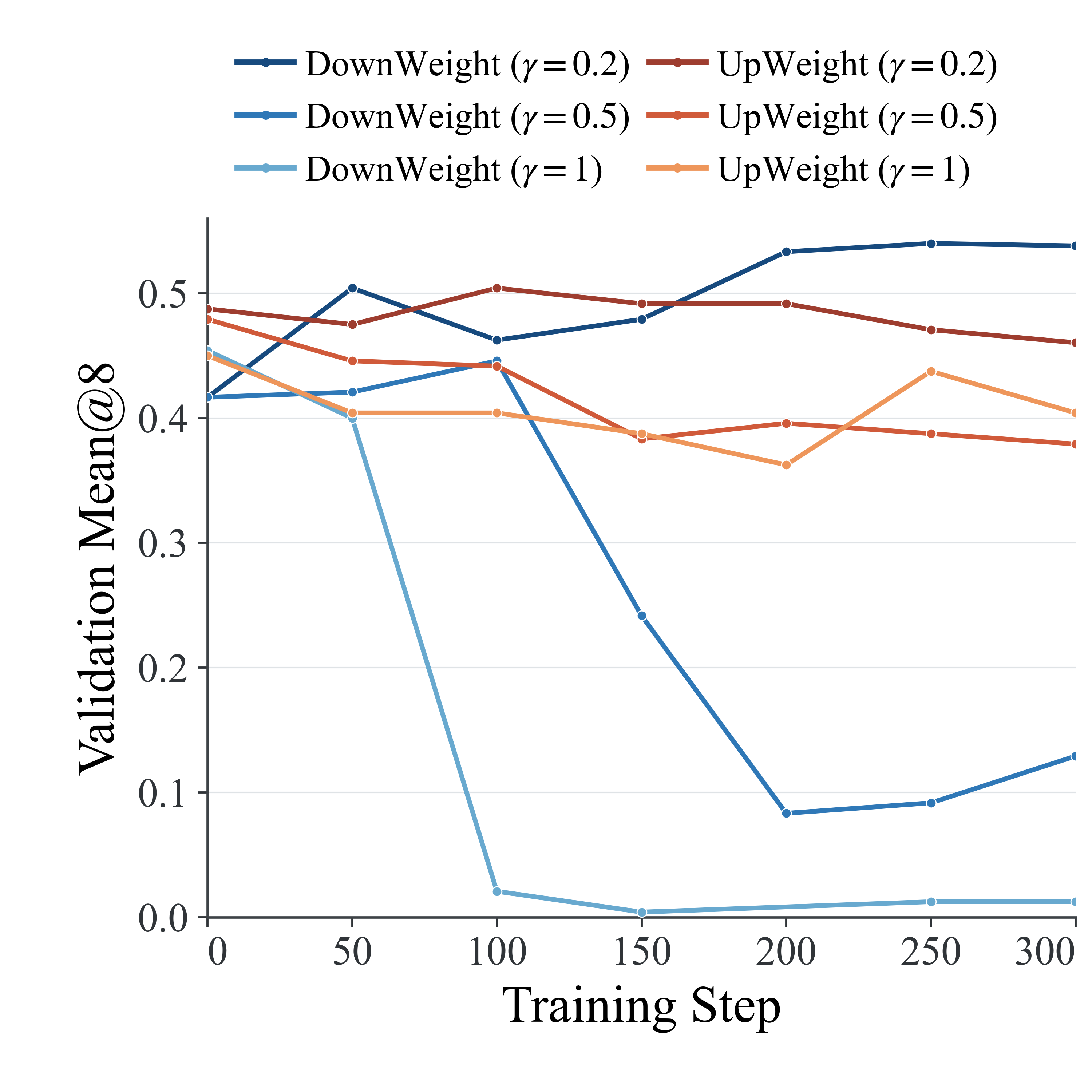}
    \caption{
    Effects of upweighting and downweighting counterfactually sensitive tokens under different modulation strengths $\gamma$.
    (a) Average reward.
    (b) Policy entropy.
    (c) Response length.
    (d) Sensitive-token fraction ($s_{i,t}\geq\lambda$).
    (e) Validation \textsc{Mean@8}.
    }
    \label{fig:cscr_gamma_ablation}
\end{figure*}

\subsection{Ablation Studies}

\paragraph{Reliability of Privileged Shift Directions.}
To test whether the sign of a privileged likelihood shift provides a reliable optimization direction, we construct Shift-Directed GRPO (SD-GRPO) as a minimal intervention on GRPO. Using the ground-truth answer as privileged information, we compute the token shift $z_{i,t}$ and assume that tokens with the largest absolute shifts may carry directional information. For tokens satisfying $|z_{i,t}|>\lambda$, we replace the verifier-derived advantage $A_i$ with $\operatorname{sign}(z_{i,t})|A_i|$, while leaving all others unchanged. We evaluate $\lambda\in\{0.05,0.1,0.2\}$ to test whether trusting a small subset of highly shifted tokens can improve optimization. Figure~\ref{fig:sd_grpo_dynamics} shows that $\lambda=0.2$, $0.1$, and $0.05$ reverse the GRPO advantage sign for approximately $1.5\%$, $3\%$, and $5\%$ of tokens, respectively. Nevertheless, all variants collapse within about 30 updates: the training reward falls from roughly $0.8$ to zero, entropy approaches zero, and response length reaches the $20{,}480$-token limit. Inspection of the resulting generations reveals persistent repetition of tokens such as \textit{the} and \textit{but}. As shown in Table~\ref{tab:shift_token_occurrences}, these highly shifted surface-form tokens receive positive shifts far more often than negative ones under privileged conditioning, causing SD-GRPO to reinforce them disproportionately. Representative cases are provided in Appendix~E. These results show that even the directions of the largest privileged shifts are unreliable for token-level credit assignment.

\paragraph{Upweighting versus Downweighting Sensitive Tokens.}
To determine whether counterfactually sensitive tokens should receive more or less credit, we compare \textsc{CSCR-UpWeight} and \textsc{CSCR-DownWeight} under the same sensitivity score, threshold, and within-trajectory normalization. For tokens above the sensitivity threshold, the former increases their raw weights toward $1+\gamma$, whereas the latter decreases them toward $1-\gamma$. We evaluate $\gamma\in\{0.2,0.5,1.0\}$ to examine both the preferred credit direction and the effect of modulation strength. Figure~\ref{fig:cscr_gamma_ablation} shows that moderate downweighting with $\gamma=0.2$ provides the most stable optimization, achieving the highest training reward, steady validation gains, gradual response-length growth, and limited entropy variation. Upweighting with $\gamma=0.2$ improves over GRPO, whereas stronger upweighting raises entropy from approximately $0.2$ to $0.6$ and $0.8$ for $\gamma=0.5$ and $1.0$, respectively, and yields little validation improvement. This suggests that mild attenuation suppresses sensitivity-driven noise while preserving enough credit on these positions to maintain stable autoregressive transitions. By contrast, downweighting with $\gamma\in\{0.5,1.0\}$ collapses: reward approaches zero, entropy spikes before collapsing, and response length falls to approximately $6{,}000$ and $2{,}000$ tokens, respectively. Under within-trajectory normalization, excessive attenuation nearly suppresses sensitive tokens while concentrating redistributed credit on the remaining positions, thereby distorting the balance required for coherent long-form reasoning. Since sensitive tokens, although less informative, still support coherent autoregressive generation, this overcorrection destabilizes optimization. Overall, sensitive tokens should be moderately downweighted rather than emphasized or nearly ignored.

\section{Conclusion}

We investigate whether privileged likelihood shifts provide reliable token-level supervision for long-CoT RLVR. Across 400 on-policy trajectories, our analysis separates shift direction from magnitude at both sampled-token and full-vocabulary levels under fixed prefixes. Counterfactual re-scoring shows that opposing outcome conditions induce predominantly same-direction shifts and substantially overlapping forward-KL signals, while large shifts concentrate on substitutable surface-form tokens rather than problem-specific reasoning content. We therefore reinterpret shift magnitude as counterfactual sensitivity and propose CSCR, a minimal GRPO extension that preserves verifier-derived direction and trajectory-level credit while reallocating credit away from highly sensitive tokens. CSCR consistently outperforms GRPO and all self-distillation baselines across both model scales and all five benchmarks. Ablations further show that trusting shift signs causes rapid optimization collapse, moderate downweighting yields the strongest and most stable gains, and stronger modulation in either direction destabilizes optimization. These results challenge privileged distributional changes as direct supervision and establish counterfactual sensitivity as a practical basis for reliable token-level credit reallocation.

\newpage

\bibliography{references}

@article{shao2024deepseekmath,
  title={Deepseekmath: Pushing the limits of mathematical reasoning in open language models},
  author={Shao, Zhihong and Wang, Peiyi and Zhu, Qihao and Xu, Runxin and Song, Junxiao and Bi, Xiao and Zhang, Haowei and Zhang, Mingchuan and Li, YK and Wu, Yang and others},
  journal={arXiv preprint arXiv:2402.03300},
  year={2024}
}

@article{guo2025deepseek,
  title={Deepseek-r1: Incentivizing reasoning capability in llms via reinforcement learning},
  author={Guo, Daya and Yang, Dejian and Zhang, Haowei and Song, Junxiao and Wang, Peiyi and Zhu, Qihao and Xu, Runxin and Zhang, Ruoyu and Ma, Shirong and Bi, Xiao and others},
  journal={arXiv preprint arXiv:2501.12948},
  year={2025}
}

@article{li2026rethinking,
  title={Rethinking on-policy distillation of large language models: Phenomenology, mechanism, and recipe},
  author={Li, Yaxuan and Zuo, Yuxin and He, Bingxiang and Zhang, Jinqian and Xiao, Chaojun and Qian, Cheng and Yu, Tianyu and Gao, Huan-ang and Yang, Wenkai and Liu, Zhiyuan and others},
  journal={arXiv preprint arXiv:2604.13016},
  year={2026}
}

@inproceedings{agarwal2024policy,
  title={On-policy distillation of language models: Learning from self-generated mistakes},
  author={Agarwal, Rishabh and Vieillard, Nino and Zhou, Yongchao and Stanczyk, Piotr and Ramos Garea, Sabela and Geist, Matthieu and Bachem, Olivier},
  booktitle={International Conference on Learning Representations},
  volume={2024},
  pages={21246--21263},
  year={2024}
}

@article{wen2025reinforcement,
  title={Reinforcement learning with verifiable rewards implicitly incentivizes correct reasoning in base llms},
  author={Wen, Xumeng and Liu, Zihan and Zheng, Shun and Ye, Shengyu and Wu, Zhirong and Wang, Yang and Xu, Zhijian and Liang, Xiao and Li, Junjie and Miao, Ziming and others},
  journal={arXiv preprint arXiv:2506.14245},
  year={2025}
}

@inproceedings{lightman2024let,
  title={Let's verify step by step},
  author={Lightman, Hunter and Kosaraju, Vineet and Burda, Yuri and Edwards, Harrison and Baker, Bowen and Lee, Teddy and Leike, Jan and Schulman, John and Sutskever, Ilya and Cobbe, Karl},
  booktitle={International Conference on Learning Representations},
  volume={2024},
  pages={39578--39601},
  year={2024}
}

@article{parthasarathi2025grpo,
  title={GRPO-$lambda $: Credit Assignment improves LLM Reasoning},
  author={Parthasarathi, Prasanna and Reymond, Mathieu and Chen, Boxing and Cui, Yufei and Chandar, Sarath},
  journal={arXiv preprint arXiv:2510.00194},
  year={2025}
}

@article{zhao2026self,
  title={Self-Distilled Reasoner: On-Policy Self-Distillation for Large Language Models},
  author={Zhao, Siyan and Xie, Zhihui and Liu, Mengchen and Huang, Jing and Pang, Guan and Chen, Feiyu and Grover, Aditya},
  journal={arXiv preprint arXiv:2601.18734},
  year={2026}
}

@article{yang2026self,
  title={Self-distilled rlvr},
  author={Yang, Chenxu and Qin, Chuanyu and Si, Qingyi and Chen, Minghui and Gu, Naibin and Yao, Dingyu and Lin, Zheng and Wang, Weiping and Wang, Jiaqi and Duan, Nan},
  journal={arXiv preprint arXiv:2604.03128},
  year={2026}
}

@article{hubotter2026reinforcement,
  title={Reinforcement Learning via Self-Distillation},
  author={H{\"u}botter, Jonas and L{\"u}beck, Frederike and Behric, Lejs and Baumann, Anton and Bagatella, Marco and Marta, Daniel and Hakimi, Ido and Shenfeld, Idan and Buening, Thomas Kleine and Guestrin, Carlos and others},
  journal={arXiv preprint arXiv:2601.20802},
  year={2026}
}

@article{li2026unifying,
  title={Unifying group-relative and self-distillation policy optimization via sample routing},
  author={Li, Gengsheng and Yang, Tianyu and Fang, Junfeng and Song, Mingyang and Zheng, Mao and Guo, Haiyun and Zhang, Dan and Wang, Jinqiao and Chua, Tat-Seng},
  journal={arXiv preprint arXiv:2604.02288},
  year={2026}
}

@article{kazemnejad2024vineppo,
  title={Vineppo: Refining credit assignment in rl training of llms},
  author={Kazemnejad, Amirhossein and Aghajohari, Milad and Portelance, Eva and Sordoni, Alessandro and Reddy, Siva and Courville, Aaron and Roux, Nicolas Le},
  journal={arXiv preprint arXiv:2410.01679},
  year={2024}
}

@article{zheng2026scope,
  title={Scope: Signal-calibrated on-policy distillation enhancement with dual-path adaptive weighting},
  author={Zheng, Binbin and Ma, Xing and Liang, Yiheng and Ruan, Jingqing and Fu, Xiaoliang and Lin, Kepeng and Zhu, Benchang and Zeng, Ke and Cai, Xunliang},
  journal={arXiv preprint arXiv:2604.10688},
  year={2026}
}

@article{yang2026prune,
  title={Prune-OPD: Efficient and Reliable On-Policy Distillation for Long-Horizon Reasoning},
  author={Yang, Zhicheng and Guo, Zhijiang and Song, Yifan and Xu, Minrui and Wang, Yongxin and Wang, Yiwei and Liang, Xiaodan and Tang, Jing},
  journal={arXiv preprint arXiv:2605.07804},
  year={2026}
}

@article{penaloza2026privileged,
  title={Privileged Information Distillation for Language Models},
  author={Penaloza, Emiliano and Vattikonda, Dheeraj and Gontier, Nicolas and Lacoste, Alexandre and Charlin, Laurent and Caccia, Massimo},
  journal={arXiv preprint arXiv:2602.04942},
  year={2026}
}

@article{pan2026rlcsd,
  title={RLCSD: Reinforcement Learning with Contrastive On-Policy Self-Distillation},
  author={Pan, Leyi and Tao, Shuchang and Zhai, Yunpeng and Zhang, Lingzhe and Liu, Zhaoyang and Ding, Bolin and Liu, Aiwei and Wen, Lijie},
  journal={arXiv preprint arXiv:2606.11709},
  year={2026}
}

@inproceedings{gu2024minillm,
  title={Minillm: Knowledge distillation of large language models},
  author={Gu, Yuxian and Dong, Li and Wei, Furu and Huang, Minlie},
  booktitle={International Conference on Learning Representations},
  volume={2024},
  pages={32694--32717},
  year={2024}
}

@article{yu2026dapo,
  title={Dapo: An open-source llm reinforcement learning system at scale},
  author={Yu, Qiying and Zhang, Zheng and Zhu, Ruofei and Yuan, Yufeng and Zuo, Xiaochen and Yue, Yu and Dai, Weinan and Fan, Tiantian and Liu, Gaohong and Liu, Lingjun and others},
  journal={Advances in Neural Information Processing Systems},
  volume={38},
  pages={113222--113244},
  year={2026}
}

@article{liu2025understanding,
  title={Understanding r1-zero-like training: A critical perspective},
  author={Liu, Zichen and Chen, Changyu and Li, Wenjun and Qi, Penghui and Pang, Tianyu and Du, Chao and Lee, Wee Sun and Lin, Min},
  journal={arXiv preprint arXiv:2503.20783},
  year={2025}
}

@misc{mathai2023amc23,
  title        = {{AMC23}: 2023 American Mathematics Competition Benchmark},
  author       = {{MAA} and {Math-AI}},
  year         = {2023},
  howpublished = {Hugging Face dataset},
  url          = {https://huggingface.co/datasets/math-ai/amc23},
  note         = {A 40-problem benchmark derived from the 2023 AMC 12A and AMC 12B competitions}
}

@misc{zhang2024aime24,
  title        = {American Invitational Mathematics Examination ({AIME}) 2024},
  author       = {Zhang, Yifan and {Math-AI}},
  year         = {2024},
  howpublished = {Hugging Face dataset},
  url          = {https://huggingface.co/datasets/math-ai/aime24}
}

@misc{zhang2025aime25,
  title        = {American Invitational Mathematics Examination ({AIME}) 2025},
  author       = {Zhang, Yifan and {Math-AI}},
  year         = {2025},
  howpublished = {Hugging Face dataset},
  url          = {https://huggingface.co/datasets/math-ai/aime25}
}

@misc{zhang2026aime26,
  title        = {American Invitational Mathematics Examination ({AIME}) 2026},
  author       = {Zhang, Yifan and {Math-AI}},
  year         = {2026},
  howpublished = {Hugging Face dataset},
  url          = {https://huggingface.co/datasets/math-ai/aime26}
}

@article{dekoninck2026matharena,
  title   = {Beyond Benchmarks: {MathArena} as an Evaluation Platform for Mathematics with {LLM}s},
  author  = {Dekoninck, Jasper and Jovanovi{\'c}, Nikola and Gehrunger, Tim
             and R{\"o}gnvalddson, K{\'a}ri and Petrov, Ivo and Sun, Chenhao
             and Vechev, Martin},
  journal = {arXiv preprint arXiv:2605.00674},
  year    = {2026},
  url     = {https://arxiv.org/abs/2605.00674}
}

@article{lambert2024tulu,
  title={Tulu 3: Pushing frontiers in open language model post-training},
  author={Lambert, Nathan and Morrison, Jacob and Pyatkin, Valentina and Huang, Shengyi and Ivison, Hamish and Brahman, Faeze and Miranda, Lester James V and Liu, Alisa and Dziri, Nouha and Lyu, Shane and others},
  journal={arXiv preprint arXiv:2411.15124},
  year={2024}
}

@article{hu2026open,
  title={Open-reasoner-zero: An open source approach to scaling up reinforcement learning on the base model},
  author={Hu, Jingcheng and Zhang, Yinmin and Han, Qi and Jiang, Daxin and Zhang, Xiangyu and Shum, Heung-Yeung},
  journal={Advances in Neural Information Processing Systems},
  volume={38},
  pages={162239--162262},
  year={2026}
}

@article{tan2025gtpo,
  title={Gtpo and grpo-s: Token and sequence-level reward shaping with policy entropy},
  author={Tan, Hongze and Wang, Zihan and Pan, Jianfei and Lin, Jinghao and Wang, Hao and Wu, Yifan and Chen, Tao and Zheng, Zhihang and Tang, Zhihao and Yang, Haihua},
  journal={arXiv preprint arXiv:2508.04349},
  year={2025}
}

@article{luo2024improve,
  title   = {Improve Mathematical Reasoning in Language Models by Automated Process Supervision},
  author  = {Luo, Liangchen and Liu, Yinxiao and Liu, Rosanne and Phatale, Samrat
             and Guo, Meiqi and Lara, Harsh and Li, Yunxuan and Shu, Lei
             and Zhu, Yun and Meng, Lei and Sun, Jiao and Rastogi, Abhinav},
  journal = {arXiv preprint arXiv:2406.06592},
  year    = {2024},
  url     = {https://arxiv.org/abs/2406.06592}
}

@article{ko2024distillm,
  title={Distillm: Towards streamlined distillation for large language models},
  author={Ko, Jongwoo and Kim, Sungnyun and Chen, Tianyi and Yun, Se-Young},
  journal={arXiv preprint arXiv:2402.03898},
  year={2024}
}

@article{jin2026entropy,
  title={Entropy-aware on-policy distillation of language models},
  author={Jin, Woogyeol and Min, Taywon and Yang, Yongjin and Kadhe, Swanand Ravindra and Zhou, Yi and Wei, Dennis and Baracaldo, Nathalie and Lee, Kimin},
  journal={arXiv preprint arXiv:2603.07079},
  year={2026}
}

@article{yang2026ogls,
  title={OGLS-SD: On-Policy Self-Distillation with Outcome-Guided Logit Steering for LLM Reasoning},
  author={Yang, Yuxiao and Wang, Xiaoyun and Zhang, Weitong},
  journal={arXiv preprint arXiv:2605.12400},
  year={2026}
}

@article{zhao2026rosd,
  title={Rosd: Reflective on-policy self-distillation for language model reasoning across domains},
  author={Zhao, Ziqi and Ma, Xinyu and Yang, Liu and Feng, Yujie and Shi, Daiting and He, Jingzhou and Xin, Xin and Ren, Zhaochun and Wu, Xiao-Ming},
  journal={arXiv preprint arXiv:2605.28014},
  year={2026}
}

@inproceedings{sheng2025hybridflow,
  title={Hybridflow: A flexible and efficient rlhf framework},
  author={Sheng, Guangming and Zhang, Chi and Ye, Zilingfeng and Wu, Xibin and Zhang, Wang and Zhang, Ru and Peng, Yanghua and Lin, Haibin and Wu, Chuan},
  booktitle={Proceedings of the Twentieth European Conference on Computer Systems},
  pages={1279--1297},
  year={2025}
}

@article{yang2025qwen3,
  title={Qwen3 technical report},
  author={Yang, An and Li, Anfeng and Yang, Baosong and Zhang, Beichen and Hui, Binyuan and Zheng, Bo and Yu, Bowen and Gao, Chang and Huang, Chengen and Lv, Chenxu and others},
  journal={arXiv preprint arXiv:2505.09388},
  year={2025}
}

\clearpage

\appendix

\section{Proof of Zero Counterfactual Perturbation Consistency for GRPO}
\label{app:grpo_zero_cpc}

We establish that, at the on-policy reference point, assigning advantages of opposite signs to the same trajectory produces antiparallel GRPO optimization signals in logit space and therefore yields zero Counterfactual Perturbation Consistency (CPC). We consider only the clipped policy-gradient term and omit the KL regularizer.

Consider a fixed problem $x$ and a trajectory
$
y_i=(y_{i,1},\ldots,y_{i,T_i})
$
sampled from the old policy $\pi_{\theta_{\mathrm{old}}}$. At token position $t$, define the policy ratio and its clipped counterpart as
\begin{equation}
r_{i,t}(\theta)
=
\frac{
\pi_\theta(y_{i,t}\mid x,y_{i,<t})
}{
\pi_{\theta_{\mathrm{old}}}(y_{i,t}\mid x,y_{i,<t})
}
\label{eq:appendix_grpo_ratio}
\end{equation}
and
\begin{equation}
\bar r_{i,t}(\theta)
=
\operatorname{clip}
\left(
r_{i,t}(\theta),
1-\epsilon,
1+\epsilon
\right).
\label{eq:appendix_grpo_clipped_ratio}
\end{equation}
For a trajectory-level advantage $A_i$, the per-token GRPO loss is
\begin{equation}
\ell_{i,t}^{\mathrm{GRPO}}(\theta;A_i)
=
-\min
\left(
r_{i,t}(\theta)A_i,
\bar r_{i,t}(\theta)A_i
\right).
\label{eq:appendix_grpo_token_loss}
\end{equation}
Throughout the derivation, $A_i$ is treated as fixed with respect to $\theta$, as in the standard GRPO policy-gradient objective.

At the on-policy reference point $\theta=\theta_{\mathrm{old}}$,
\begin{equation}
r_{i,t}(\theta_{\mathrm{old}})=1.
\label{eq:appendix_ratio_at_old_policy}
\end{equation}
Because $1$ lies strictly inside the clipping interval, continuity of $r_{i,t}(\theta)$ implies that clipping is inactive in a neighborhood of $\theta_{\mathrm{old}}$. Within this neighborhood,
\begin{equation}
\bar r_{i,t}(\theta)=r_{i,t}(\theta),
\label{eq:appendix_clipping_inactive}
\end{equation}
and the local loss reduces to
\begin{equation}
\ell_{i,t}^{\mathrm{GRPO}}(\theta;A_i)
=
-A_i r_{i,t}(\theta).
\label{eq:appendix_unclipped_local_loss}
\end{equation}

Let $\mathbf a_{i,t}$ denote the current-policy logits at position $t$, and let
\begin{equation}
p_{i,t}(v)
=
\pi_\theta(v\mid x,y_{i,<t})
=
\operatorname{softmax}(\mathbf a_{i,t})_v,
\qquad
v\in\mathcal V.
\label{eq:appendix_current_distribution}
\end{equation}
The policy ratio can then be written as
\begin{equation}
r_{i,t}(\theta)
=
\frac{
p_{i,t}(y_{i,t})
}{
p_{i,t}^{\mathrm{old}}(y_{i,t})
},
\label{eq:appendix_ratio_distribution_form}
\end{equation}
where
$
p_{i,t}^{\mathrm{old}}(v)
=
\pi_{\theta_{\mathrm{old}}}(v\mid x,y_{i,<t})
$
is independent of the current logits. Applying the standard softmax derivative gives, for any $v\in\mathcal V$,
\begin{align}
\frac{\partial r_{i,t}(\theta)}
{\partial a_{i,t}(v)}
&=
\frac{1}
{p_{i,t}^{\mathrm{old}}(y_{i,t})}
\frac{\partial p_{i,t}(y_{i,t})}
{\partial a_{i,t}(v)}
\nonumber\\
&=
r_{i,t}(\theta)
\left(
\mathbb{I}[v=y_{i,t}]
-
p_{i,t}(v)
\right).
\label{eq:appendix_ratio_logit_derivative}
\end{align}
Equivalently,
\begin{equation}
\nabla_{\mathbf a_{i,t}}r_{i,t}(\theta)
=
r_{i,t}(\theta)
\left(
\mathbf e_{y_{i,t}}
-
\mathbf p_{i,t}
\right),
\label{eq:appendix_ratio_logit_gradient}
\end{equation}
where $\mathbf e_{y_{i,t}}$ is the one-hot vector associated with the sampled token and $\mathbf p_{i,t}$ is the full-vocabulary current-policy distribution.

Combining Eqs.~\ref{eq:appendix_unclipped_local_loss} and~\ref{eq:appendix_ratio_logit_gradient}, the logit-space gradient of the GRPO loss is
\begin{equation}
\nabla_{\mathbf a_{i,t}}
\ell_{i,t}^{\mathrm{GRPO}}(\theta;A_i)
=
-A_i r_{i,t}(\theta)
\left(
\mathbf e_{y_{i,t}}
-
\mathbf p_{i,t}
\right).
\label{eq:appendix_grpo_logit_gradient}
\end{equation}
Define the corresponding gradient-descent signal as
\begin{equation}
\mathbf g_{i,t}(A_i)
=
-
\nabla_{\mathbf a_{i,t}}
\ell_{i,t}^{\mathrm{GRPO}}(\theta;A_i).
\label{eq:appendix_grpo_descent_signal_definition}
\end{equation}
It follows that
\begin{equation}
\mathbf g_{i,t}(A_i)
=
A_i r_{i,t}(\theta)
\left(
\mathbf e_{y_{i,t}}
-
\mathbf p_{i,t}
\right).
\label{eq:appendix_grpo_descent_signal}
\end{equation}
At $\theta=\theta_{\mathrm{old}}$, this reduces to
\begin{equation}
\mathbf g_{i,t}(A_i)
=
A_i
\left(
\mathbf e_{y_{i,t}}
-
\mathbf p_{i,t}^{\mathrm{old}}
\right).
\label{eq:appendix_grpo_signal_at_old_policy}
\end{equation}

Now consider two counterfactual assignments of opposite-sign advantages to the same fixed trajectory:
\begin{equation}
A_i^{+}>0,
\qquad
A_i^{-}<0.
\label{eq:appendix_opposite_advantages}
\end{equation}
At the on-policy reference point, or more generally within any region where clipping is inactive, both assignments share the same problem, trajectory, prefix, sampled token, probability ratio, and policy distribution. Their gradient-descent signals therefore satisfy
\begin{align}
\mathbf g_{i,t}^{+}
&=
A_i^{+}r_{i,t}
\left(
\mathbf e_{y_{i,t}}
-
\mathbf p_{i,t}
\right),
\label{eq:appendix_positive_grpo_signal}\\
\mathbf g_{i,t}^{-}
&=
A_i^{-}r_{i,t}
\left(
\mathbf e_{y_{i,t}}
-
\mathbf p_{i,t}
\right).
\label{eq:appendix_negative_grpo_signal}
\end{align}
Therefore,
\begin{equation}
\mathbf g_{i,t}^{-}
=
-
\frac{|A_i^{-}|}{A_i^{+}}
\mathbf g_{i,t}^{+}.
\label{eq:appendix_antiparallel_signals}
\end{equation}
Thus, the two signals are antiparallel at every token position. This result requires only that the advantages have opposite signs; their magnitudes need not be equal.

Following the CPC definition in the main text, define the trajectory-level combined magnitude and disagreement magnitude as
\begin{align}
\mathcal M_{\mathrm{GRPO}}
&=
\frac{1}{T_i}
\sum_{t=1}^{T_i}
\left(
\left\|\mathbf g_{i,t}^{+}\right\|_1
+
\left\|\mathbf g_{i,t}^{-}\right\|_1
\right),
\label{eq:appendix_grpo_combined_magnitude}\\
\mathcal D_{\mathrm{GRPO}}
&=
\frac{1}{T_i}
\sum_{t=1}^{T_i}
\left\|
\mathbf g_{i,t}^{+}
-
\mathbf g_{i,t}^{-}
\right\|_1.
\label{eq:appendix_grpo_disagreement}
\end{align}
Using Eq.~\ref{eq:appendix_antiparallel_signals}, we obtain
\begin{align}
\left\|
\mathbf g_{i,t}^{+}
-
\mathbf g_{i,t}^{-}
\right\|_1
&=
\left\|
\left(
1+\frac{|A_i^{-}|}{A_i^{+}}
\right)
\mathbf g_{i,t}^{+}
\right\|_1
\nonumber\\
&=
\left\|\mathbf g_{i,t}^{+}\right\|_1
+
\left\|\mathbf g_{i,t}^{-}\right\|_1.
\label{eq:appendix_grpo_no_overlap}
\end{align}
Consequently,
\begin{equation}
\mathcal D_{\mathrm{GRPO}}
=
\mathcal M_{\mathrm{GRPO}}.
\label{eq:appendix_grpo_equal_magnitude_disagreement}
\end{equation}
For $\mathcal M_{\mathrm{GRPO}}>0$, substitution into the CPC definition yields
\begin{equation}
\mathrm{CPC}_{\mathrm{GRPO}}
=
1-
\frac{
\mathcal D_{\mathrm{GRPO}}
}{
\mathcal M_{\mathrm{GRPO}}
}
=
0.
\label{eq:appendix_grpo_zero_cpc}
\end{equation}

Therefore, at the on-policy reference point, counterfactually assigning opposite-sign GRPO advantages to the same trajectory produces antiparallel logit-space optimization signals with no same-coordinate, same-direction overlap. The same conclusion holds more generally whenever PPO clipping is inactive. This establishes the zero-overlap reference used to evaluate privileged forward-KL signals in the main text.

\section{Top-50 Tokens under Counterfactual Privileged Prompts}
\label{app:top50_tokens_counterfactual_prompts}

\begin{table*}[!t]
\caption{
Top-50 most frequent sampled tokens with large positive, large negative, or near-zero privileged likelihood shifts across 400 on-policy trajectories. Only results for $\lambda=0.1$ are reported. Numbers in parentheses denote occurrence counts.
}
\label{tab:appendix_top50_shift_tokens}
\centering
\scriptsize
\setlength{\tabcolsep}{1.7pt}
\renewcommand{\arraystretch}{0.82}

\newcommand{\tokb}[2]{\texttt{#1}\,(#2)}

\begin{tabular*}{\textwidth}{
@{\extracolsep{\fill}}
r
llllll
@{}
}
\toprule
&
\multicolumn{2}{c}{\textbf{Large Positive Shifts}}
&
\multicolumn{2}{c}{\textbf{Large Negative Shifts}}
&
\multicolumn{2}{c}{\textbf{Near-Zero Shifts}}
\\
\cmidrule(lr){2-3}
\cmidrule(lr){4-5}
\cmidrule(lr){6-7}
\textbf{Rank}
&
$\boldsymbol{z^{\mathrm{pos}}>0.1}$
&
$\boldsymbol{z^{\mathrm{neg}}>0.1}$
&
$\boldsymbol{z^{\mathrm{pos}}<-0.1}$
&
$\boldsymbol{z^{\mathrm{neg}}<-0.1}$
&
$\boldsymbol{|z^{\mathrm{pos}}|<0.01}$
&
$\boldsymbol{|z^{\mathrm{neg}}|<0.01}$
\\
\midrule

1
& \tokb{The}{8328}
& \tokb{The}{13979}
& \tokb{.}{6994}
& \tokb{.}{9576}
& \tokb{<Whitespace>}{256692}
& \tokb{<Whitespace>}{250467}
\\

2
& \tokb{.}{6233}
& \tokb{.}{11061}
& \tokb{The}{4652}
& \tokb{Therefore}{6028}
& \tokb{,}{168425}
& \tokb{,}{159943}
\\

3
& \tokb{,}{4354}
& \tokb{,}{7399}
& \tokb{Therefore}{3566}
& \tokb{The}{5530}
& \tokb{1}{152042}
& \tokb{1}{151250}
\\

4
& \tokb{But}{3855}
& \tokb{So}{6715}
& \tokb{So}{3488}
& \tokb{Is}{5170}
& \tokb{2}{145113}
& \tokb{2}{144428}
\\

5
& \tokb{<Whitespace>}{3684}
& \tokb{But}{6248}
& \tokb{Is}{3390}
& \tokb{,}{4227}
& \tokb{0}{110276}
& \tokb{0}{109945}
\\

6
& \tokb{So}{3400}
& \tokb{<Whitespace>}{5792}
& \tokb{,}{3174}
& \tokb{So}{4184}
& \tokb{=}{90223}
& \tokb{=}{87722}
\\

7
& \tokb{Let}{2917}
& \tokb{Let}{4027}
& \tokb{<Whitespace>}{3076}
& \tokb{<Whitespace>}{3872}
& \tokb{.}{90080}
& \tokb{3}{81732}
\\

8
& \tokb{This}{2012}
& \tokb{This}{3463}
& \tokb{Let}{2398}
& \tokb{Wait}{3733}
& \tokb{3}{82219}
& \tokb{.}{76677}
\\

9
& \tokb{A}{1706}
& \tokb{For}{3188}
& \tokb{But}{2387}
& \tokb{Let}{3721}
& \tokb{The}{66152}
& \tokb{5}{60815}
\\

10
& \tokb{For}{1686}
& \tokb{'s}{2756}
& \tokb{Wait}{2041}
& \tokb{But}{3158}
& \tokb{5}{61066}
& \tokb{+}{56800}
\\

11
& \tokb{And}{1590}
& \tokb{A}{2747}
& \tokb{:}{1848}
& \tokb{?}{2766}
& \tokb{+}{57364}
& \tokb{The}{56605}
\\

12
& \tokb{=}{1589}
& \tokb{=}{2456}
& \tokb{I}{1799}
& \tokb{:}{2724}
& \tokb{Is}{56566}
& \tokb{4}{55509}
\\

13
& \tokb{Is}{1533}
& \tokb{(}{2433}
& \tokb{That}{1761}
& \tokb{Maybe}{2531}
& \tokb{4}{55806}
& \tokb{(}{53315}
\\

14
& \tokb{(}{1517}
& \tokb{And}{2368}
& \tokb{Maybe}{1697}
& \tokb{Alternatively}{2233}
& \tokb{(}{55580}
& \tokb{-}{52511}
\\

15
& \tokb{Wait}{1498}
& \tokb{Is}{2255}
& \tokb{?}{1690}
& \tokb{That}{2227}
& \tokb{-}{53237}
& \tokb{Is}{51896}
\\

16
& \tokb{'s}{1476}
& \tokb{If}{2233}
& \tokb{If}{1561}
& \tokb{If}{2173}
& \tokb{6}{45496}
& \tokb{6}{45289}
\\

17
& \tokb{If}{1326}
& \tokb{It}{2184}
& \tokb{We}{1362}
& \tokb{We}{2081}
& \tokb{)}{42552}
& \tokb{)}{41158}
\\

18
& \tokb{There}{1260}
& \tokb{:}{2060}
& \tokb{Alternatively}{1349}
& \tokb{Me}{2053}
& \tokb{/}{36919}
& \tokb{/}{36099}
\\

19
& \tokb{:}{1259}
& \tokb{That}{1902}
& \tokb{Since}{1188}
& \tokb{Since}{1914}
& \tokb{A}{35137}
& \tokb{A}{32937}
\\

20
& \tokb{It}{1224}
& \tokb{In}{1751}
& \tokb{(}{1178}
& \tokb{I}{1888}
& \tokb{Of}{32572}
& \tokb{Of}{31157}
\\

21
& \tokb{In}{1127}
& \tokb{There}{1663}
& \tokb{For}{1160}
& \tokb{Then}{1881}
& \tokb{<U+FFFD>}{30943}
& \tokb{<U+FFFD>}{30519}
\\

22
& \tokb{That}{1090}
& \tokb{Wait}{1654}
& \tokb{A}{1128}
& \tokb{(}{1369}
& \tokb{:}{27543}
& \tokb{8}{26287}
\\

23
& \tokb{We}{937}
& \tokb{No}{1486}
& \tokb{Then}{1098}
& \tokb{A}{1368}
& \tokb{8}{26418}
& \tokb{9}{25823}
\\

24
& \tokb{1}{793}
& \tokb{Each}{1317}
& \tokb{In}{971}
& \tokb{And}{1302}
& \tokb{9}{25959}
& \tokb{:}{25333}
\\

25
& \tokb{Therefore}{772}
& \tokb{I}{1215}
& \tokb{And}{958}
& \tokb{Which}{1261}
& \tokb{7}{25319}
& \tokb{7}{25181}
\\

26
& \tokb{Maybe}{767}
& \tokb{We}{1171}
& \tokb{Me}{931}
& \tokb{However}{1245}
& \tokb{To}{25310}
& \tokb{To}{24209}
\\

27
& \tokb{Each}{764}
& \tokb{To}{1166}
& \tokb{Which}{918}
& \tokb{In}{1224}
& \tokb{And}{20196}
& \tokb{And}{18271}
\\

28
& \tokb{To}{759}
& \tokb{Are}{1153}
& \tokb{Of}{906}
& \tokb{Think}{1206}
& \tokb{*}{18507}
& \tokb{*}{17774}
\\

29
& \tokb{Then}{735}
& \tokb{1}{1062}
& \tokb{Check}{818}
& \tokb{).}{1174}
& \tokb{N}{18309}
& \tokb{N}{17695}
\\

30
& \tokb{No}{727}
& \tokb{Maybe}{1054}
& \tokb{1}{798}
& \tokb{Of}{1153}
& \tokb{\$}{17778}
& \tokb{\$}{17410}
\\

31
& \tokb{Since}{717}
& \tokb{Would}{1048}
& \tokb{Problem}{756}
& \tokb{For}{1149}
& \tokb{B}{17507}
& \tokb{B}{17268}
\\

32
& \tokb{Of}{681}
& \tokb{Then}{1016}
& \tokb{However}{749}
& \tokb{With}{1127}
& \tokb{X}{17178}
& \tokb{X}{16713}
\\

33
& \tokb{I}{648}
& \tokb{)}{937}
& \tokb{With}{724}
& \tokb{Check}{1020}
& \tokb{\textbackslash}{16477}
& \tokb{\textbackslash}{16369}
\\

34
& \tokb{Are}{626}
& \tokb{Of}{915}
& \tokb{First}{681}
& \tokb{Compute}{972}
& \tokb{\textsuperscript{2}}{14507}
& \tokb{\textsuperscript{2}}{14383}
\\

35
& \tokb{)}{569}
& \tokb{Since}{847}
& \tokb{=}{656}
& \tokb{=}{943}
& \tokb{K}{14260}
& \tokb{K}{13834}
\\

36
& \tokb{-}{565}
& \tokb{),}{825}
& \tokb{Not}{646}
& \tokb{1}{935}
& \tokb{Are}{13238}
& \tokb{\}}{12696}
\\

37
& \tokb{From}{549}
& \tokb{From}{819}
& \tokb{).}{635}
& \tokb{Can}{929}
& \tokb{\}}{12803}
& \tokb{Are}{12054}
\\

38
& \tokb{With}{529}
& \tokb{-}{802}
& \tokb{There}{624}
& \tokb{There}{928}
& \tokb{So}{12780}
& \tokb{Y}{11886}
\\

39
& \tokb{2}{523}
& \tokb{Therefore}{793}
& \tokb{2}{612}
& \tokb{Not}{901}
& \tokb{That}{12259}
& \tokb{C}{11804}
\\

40
& \tokb{Would}{519}
& \tokb{Check}{778}
& \tokb{Number}{582}
& \tokb{First}{888}
& \tokb{Y}{12089}
& \tokb{\ensuremath{\sqrt{\phantom{0}}}}{11006}
\\

41
& \tokb{Total}{510}
& \tokb{Total}{761}
& \tokb{To}{571}
& \tokb{Problem}{888}
& \tokb{C}{12051}
& \tokb{Be}{10747}
\\

42
& \tokb{Now}{487}
& \tokb{Where}{740}
& \tokb{Can}{570}
& \tokb{-}{767}
& \tokb{Me}{11708}
& \tokb{That}{10676}
\\

43
& \tokb{N}{470}
& \tokb{Which}{725}
& \tokb{Compute}{562}
& \tokb{Number}{767}
& \tokb{Be}{11095}
& \tokb{\{}{10581}
\\

44
& \tokb{Which}{468}
& \tokb{Now}{713}
& \tokb{Think}{546}
& \tokb{2}{746}
& \tokb{\ensuremath{\sqrt{\phantom{0}}}}{11060}
& \tokb{\textasciicircum}{9838}
\\

45
& \tokb{?}{457}
& \tokb{At}{712}
& \tokb{-}{524}
& \tokb{Another}{722}
& \tokb{\{}{10595}
& \tokb{Me}{9592}
\\

46
& \tokb{Think}{456}
& \tokb{2}{705}
& \tokb{)}{474}
& \tokb{To}{708}
& \tokb{But}{10247}
& \tokb{M}{9358}
\\

47
& \tokb{X}{453}
& \tokb{X}{705}
& \tokb{Answer}{470}
& \tokb{Need}{698}
& \tokb{\textasciicircum}{10033}
& \tokb{So}{8975}
\\

48
& \tokb{),}{451}
& \tokb{With}{693}
& \tokb{Each}{463}
& \tokb{Numbers}{692}
& \tokb{M}{9688}
& \tokb{R}{8787}
\\

49
& \tokb{3}{439}
& \tokb{N}{641}
& \tokb{Have}{447}
& \tokb{/}{666}
& \tokb{In}{9304}
& \tokb{In}{8113}
\\

50
& \tokb{At}{410}
& \tokb{*}{638}
& \tokb{Numbers}{444}
& \tokb{)}{654}
& \tokb{R}{9029}
& \tokb{P}{7913}
\\

\bottomrule
\end{tabular*}
\end{table*}

\paragraph{Large Shifts Predominantly Affect Surface-Form Tokens.}
Table~\ref{tab:appendix_top50_shift_tokens} shows a consistent separation between tokens with large and near-zero likelihood shifts. Under both counterfactual privileged prompts, the large-shift lists are dominated by discourse markers, local transition cues, and other highly substitutable surface-form tokens, including \textit{But}, \textit{So}, \textit{Let}, \textit{Wait}, \textit{Therefore}, and \textit{Maybe}. By contrast, the near-zero lists are enriched with digits, variables, arithmetic operators, and mathematical notation, such as \texttt{0--9}, \texttt{=}, \texttt{+}, \texttt{-}, \texttt{/}, and \texttt{\ensuremath{\sqrt{\phantom{0}}}}. Although neither category is semantically exclusive, the contrast is stable under both $c^{\mathrm{pos}}$ and $c^{\mathrm{neg}}$: privileged conditioning primarily perturbs stylistic realization and discourse organization, while tokens encoding problem-specific calculations are substantially less sensitive.

\paragraph{Sensitive Token Identities Are Stable across Counterfactual Prompts.}
The tokens most sensitive to the positive and negative privileged prompts exhibit substantial overlap. Among the Top-50 tokens with positive likelihood shifts, 47 appear under both $z^{\mathrm{pos}}>0.1$ and $z^{\mathrm{neg}}>0.1$, corresponding to a Jaccard similarity of $47/53=88.7\%$. The Top-50 negative-shift lists exhibit the same overlap, sharing 47 tokens with a Jaccard similarity of $88.7\%$. When the positive and negative tails are combined, the two privileged conditions contain 64 and 66 unique high-shift tokens, respectively, of which 61 are shared, yielding a Jaccard similarity of $88.4\%$. The near-zero lists are even more consistent, sharing 49 of their Top-50 entries. These results indicate that reversing the semantic polarity of privileged conditioning changes the magnitude and ranking of individual shifts more than it changes which token types are sensitive.

\paragraph{Positive and Negative Shifts Affect Largely Overlapping Token Classes.}
Positive and negative likelihood shifts also affect broadly similar token classes within each privileged condition. Under $c^{\mathrm{pos}}$, the Top-50 positive- and negative-shift lists account for 71,445 and 72,021 sampled-token occurrences, respectively, and share 36 token identities. Under $c^{\mathrm{neg}}$, the corresponding lists account for 115,771 and 102,072 occurrences and share 34 token identities. Thus, neither shift direction is confined to a distinct semantic class: many of the same discourse markers and highly substitutable surface-form tokens occur frequently in both tails, with their signs varying across contexts and token positions. Although the relative frequencies differ more under $c^{\mathrm{neg}}$, the substantial identity overlap remains stable across both conditions. Consequently, shift sign provides limited semantic separation, whereas shift magnitude more consistently identifies general sensitivity to privileged conditioning.

\section{Robustness across Alternative Counterfactual Privileged Prompts}

\begin{table*}[!t]
\caption{
Alternative counterfactual privileged prompt constructions.
The fixed sampled trajectory is inserted as
\texttt{\{current\_solution\}} and evaluated under teacher forcing.
}
\label{tab:alternative_counterfactual_prompts}
\centering
\footnotesize
\setlength{\tabcolsep}{5pt}
\renewcommand{\arraystretch}{1.08}

\begin{tabularx}{\textwidth}{
    @{}p{0.18\textwidth}
    >{\raggedright\arraybackslash}X@{}
}
\toprule
\textbf{Construction} & \textbf{Prompt Content} \\
\midrule

\textbf{Shared Instruction}
&
You are a helpful math assistant.

Please solve the math problem step by step clearly and concisely.

You must enclose your final answer exactly within
\texttt{\string\boxed\{\{\}\}}.

\texttt{\{question\}}
\\

\midrule

\textbf{Answer-Level Counterfactual Feedback}
&
\textbf{Positive condition}

The candidate final answer
\texttt{\string\boxed\{\{answer\}\}}
has been marked as correct by a verifier.
With this feedback in mind, solve the problem step by step.

\texttt{\{current\_solution\}}

\par\medskip

\textbf{Negative condition}

The candidate final answer
\texttt{\string\boxed\{\{answer\}\}}
has been marked as incorrect by a verifier.
With this feedback in mind, solve the problem step by step.

\texttt{\{current\_solution\}}

\par\medskip

\textit{Counterfactual instantiation.}
In both conditions, \texttt{\{answer\}} is instantiated with the same
ground-truth answer. The positive condition affirms this answer, whereas
the negative condition counterfactually marks the same answer as incorrect.
\\

\midrule

\textbf{Mild Outcome-Only Counterfactual Feedback}
&
\textbf{Positive condition}

A gold-standard final-answer verifier has judged that the assistant's
following solution reaches the correct final answer.

This is outcome-level feedback only: it does not provide a reference
solution, intermediate reasoning steps, or the final answer.

After considering this outcome feedback, follow the solution trajectory
step by step.
Strengthen the key assumptions, transformations, calculations, and
intermediate conclusions that support the verified correct final-answer
outcome.

\texttt{\{current\_solution\}}

\par\medskip

\textbf{Negative condition}

A gold-standard final-answer verifier has judged that the assistant's
following solution does not reach the correct final answer.

This is outcome-level feedback only: it does not provide a reference
solution, identify the error location, or reveal the correct final answer.

After considering this outcome feedback, follow the solution trajectory
step by step.
Pay special attention to assumptions, transformations, shortcuts,
calculations, and intermediate conclusions that may lead to the incorrect
final-answer outcome.
Weaken unreliable reasoning paths, revise them when necessary, and
backtrack or reconsider if needed.

\texttt{\{current\_solution\}}

\par\medskip

\textit{Counterfactual instantiation.}
The two conditions differ only in the asserted trajectory-level verifier
outcome. Neither condition provides the ground-truth answer, a reference
solution, or explicit process-level supervision.
\\

\midrule

\textbf{Reference-Solution Counterfactual Conditioning}
&
The following is a useful reference solution:

\texttt{\{reference\_solution\}}

After reading the reference solution, solve the problem step by step.

\texttt{\{current\_solution\}}

\par\medskip

\textit{Counterfactual instantiation.}
Under the positive condition,
\texttt{\{reference\_solution\}} is instantiated with a correct reasoning
process. Under the negative condition, it is instantiated with an incorrect
reasoning process. The paired solutions are generated by DeepSeek-V4-Pro
and manually reviewed to verify their intended correctness polarity.
\\

\bottomrule
\end{tabularx}
\end{table*}

\subsection{Alternative Counterfactual Privileged Prompt Constructions}
\label{app:alternative_counterfactual_prompts}

To assess whether our findings depend on the deliberately polarized prompts used in the main text, we construct three additional pairs of counterfactual privileged prompts, summarized in Table~\ref{tab:alternative_counterfactual_prompts}. For all constructions, we keep the problem, sampled trajectory, and every token prefix fixed, modifying only the additional privileged context used during likelihood re-scoring. Within each construction, the positive and negative conditions follow the same prompt template and differ only in the privileged information that determines their intended correctness polarity.

The \emph{answer-level} construction holds the ground-truth answer fixed and counterfactually describes the same answer as either correct or incorrect. It therefore isolates the effect of asserted answer-level correctness while controlling for the answer content itself. The \emph{mild outcome-only} construction states only whether the sampled trajectory reaches the correct final answer, without revealing the answer, providing a reference solution, identifying a specific error, or explicitly evaluating the quality of the reasoning process. Compared with the polarized prompts used in the main text, this construction removes strongly evaluative language and retains only concise trajectory-level outcome feedback. Finally, the \emph{reference-solution} construction provides either a correct or an incorrect reasoning process through otherwise identical prompt templates. The correct and incorrect reference solutions are generated by DeepSeek-V4-Pro and subsequently reviewed manually to confirm that they satisfy their designated correctness labels.

Together, these three constructions cover privileged information at the answer, trajectory-outcome, and full-reasoning levels. They allow us to determine whether the observed likelihood-shift patterns persist across substantially different forms and strengths of counterfactual conditioning, rather than arising specifically from the polarized wording adopted in the main experiments.

\subsection{Diagnostic Results Across Alternative Prompt Constructions}
\label{app:alternative_prompt_results}

We repeat the diagnostic analyses using the three alternative prompt pairs introduced in Appendix~\ref{app:alternative_counterfactual_prompts}. All experiments use Qwen3-1.7B and the same 400 fixed on-policy trajectories as in the main analysis. The problem, sampled trajectory, token prefixes, and unprivileged distributions are held fixed; only the counterfactual privileged prompt is changed. We first examine whether the full-vocabulary probability shifts induced by each prompt pair exhibit the outcome-aligned directional reversal expected from opposing verifier semantics.

\begin{figure*}[t]
    \centering
    \begin{minipage}[t]{0.325\textwidth}
        \centering
        \includegraphics[width=\linewidth]
        {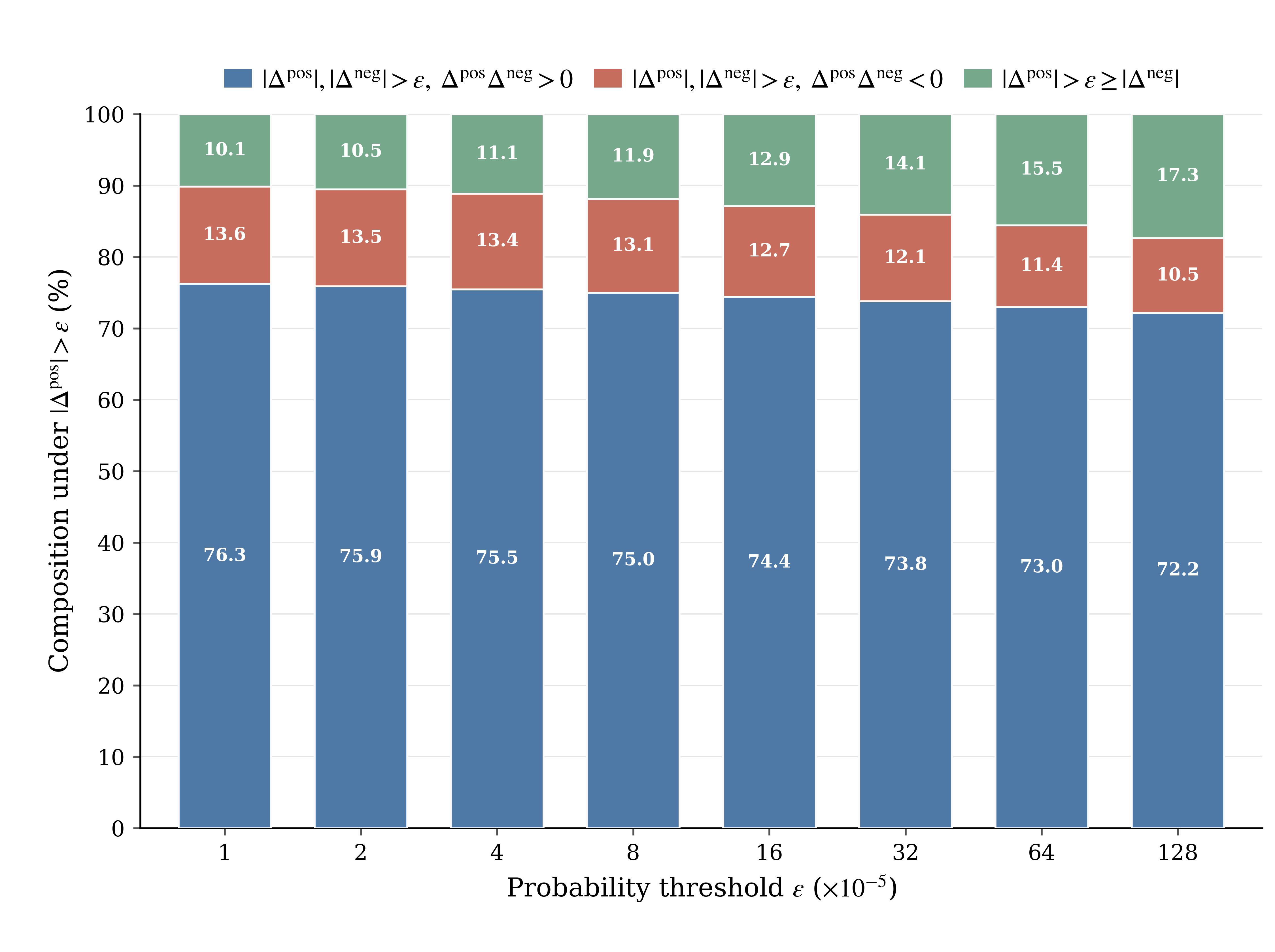}

        {\small (a) Answer-level feedback.}
    \end{minipage}
    \hfill
    \begin{minipage}[t]{0.325\textwidth}
        \centering
        \includegraphics[width=\linewidth]
        {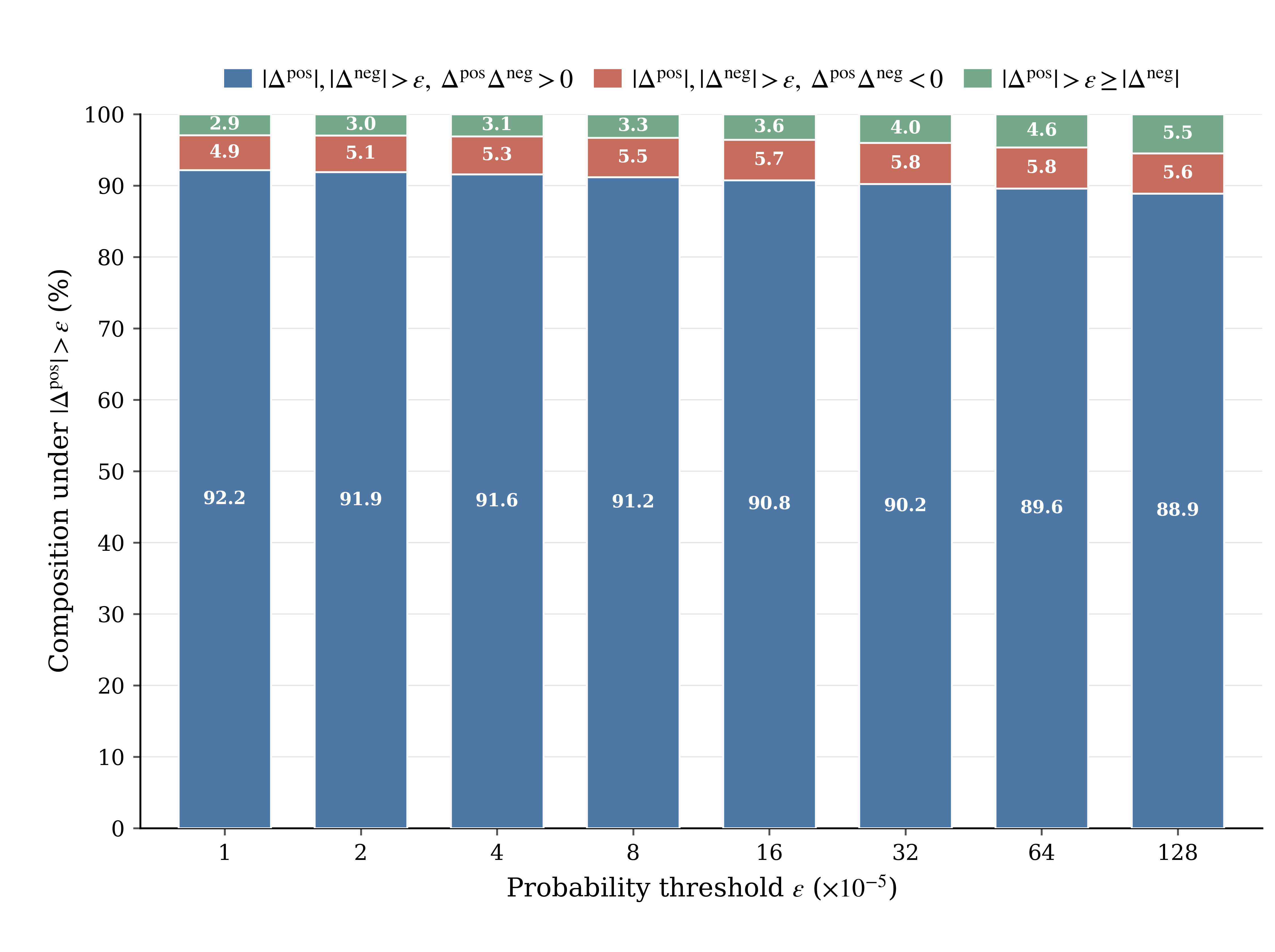}

        {\small (b) Mild outcome-only feedback.}
    \end{minipage}
    \hfill
    \begin{minipage}[t]{0.325\textwidth}
        \centering
        \includegraphics[width=\linewidth]
        {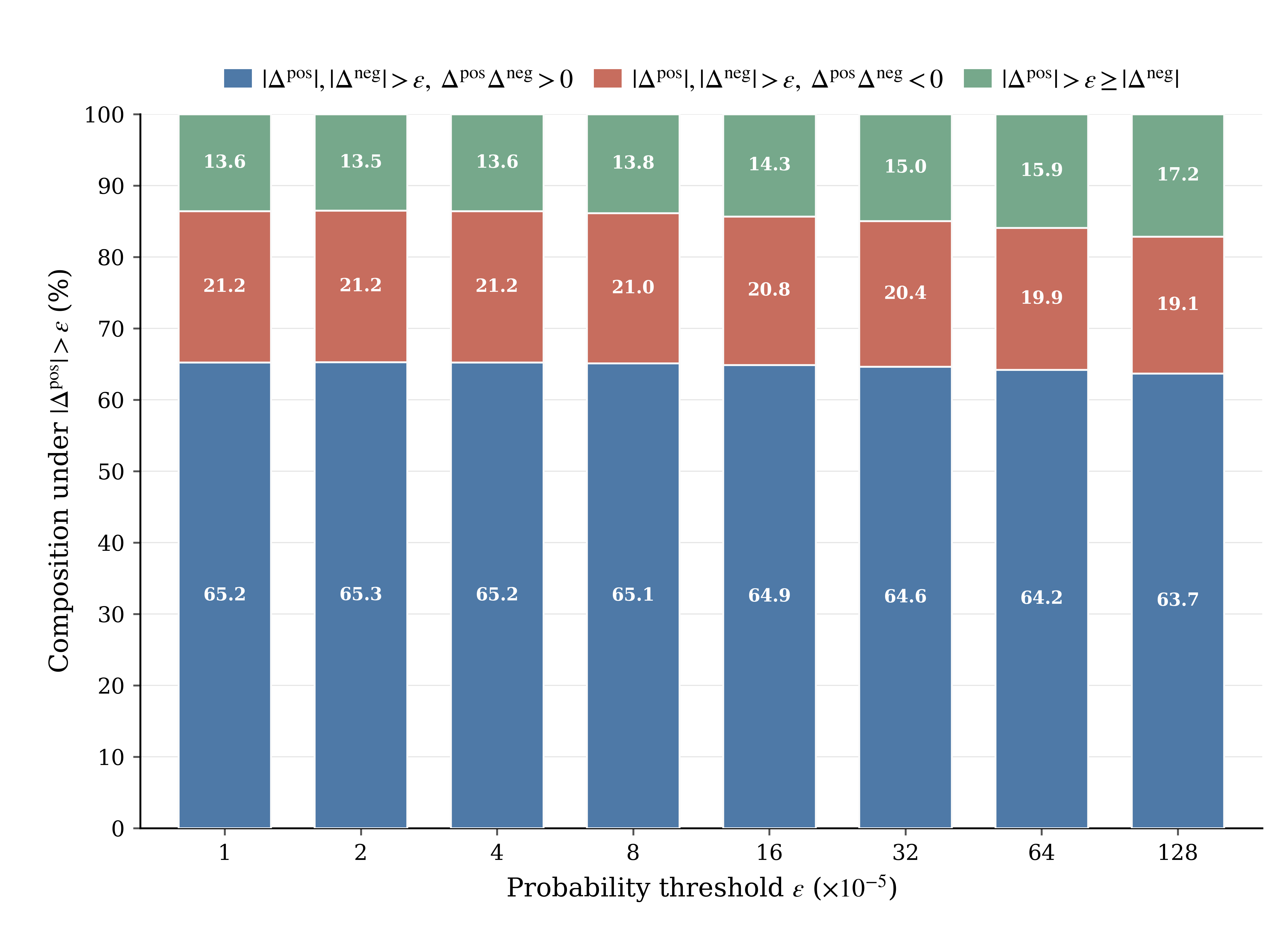}

        {\small (c) Reference-solution conditioning.}
    \end{minipage}

    \caption{
    Full-vocabulary probability-shift composition under three alternative counterfactual privileged prompt pairs. For each threshold $\epsilon$, entries satisfying
    $|\Delta_i^{\mathrm{pos}}(v)|>\epsilon$ form the reference set and are partitioned into jointly significant same-sign shifts, jointly significant opposite-sign shifts, and shifts significant only under the positive condition. Results are obtained from Qwen3-1.7B on the same 400 fixed on-policy trajectories used in the main analysis.
    }
    \label{fig:alternative_full_vocab_shift_composition}
\end{figure*}

\paragraph{Full-Vocabulary Shift Composition.}
Figure~\ref{fig:alternative_full_vocab_shift_composition} shows that same-sign shifts remain the dominant component under all three alternative prompt constructions and across all probability thresholds. Under answer-level counterfactual feedback, jointly significant same-sign shifts account for $72.2\%$--$76.3\%$ of the entries significant under the positive privileged condition, whereas opposite-sign shifts account for only $10.5\%$--$13.6\%$. The dominance is even stronger under mild outcome-only feedback: same-sign shifts constitute $88.9\%$--$92.2\%$ of the reference set, while only $4.9\%$--$5.8\%$ reverse direction. Thus, removing the strongly evaluative praise and criticism used in the main-text prompts does not reduce coordinate-wise directional agreement. Instead, the two opposing outcome conditions produce an even larger proportion of probability shifts pointing in the same direction.

Reference-solution conditioning introduces the greatest content-level difference between the two privileged conditions, because the model receives a complete correct or incorrect reasoning process. This construction produces a larger proportion of opposite-sign shifts than the other alternatives, but same-sign changes nevertheless remain predominant: they account for $63.7\%$--$65.3\%$ of the reference set, compared with $19.1\%$--$21.2\%$ for opposite-sign changes. Even complete reasoning processes with opposing correctness polarity therefore fail to induce a predominantly reversed coordinate-wise shift pattern.

As $\epsilon$ increases, the same-sign proportion varies only moderately, while the fraction significant exclusively under the positive condition generally increases. Crucially, the opposite-sign proportion remains small or decreases rather than becoming dominant. The observed directional agreement is therefore not confined to weak perturbations near zero. Across answer-level judgments, mild trajectory-level outcomes, and complete reference solutions, opposing privileged semantics primarily preserve common full-vocabulary update directions instead of reversing them. These results confirm that the main-text diagnosis is robust to prompt construction and further support interpreting privileged likelihood shifts as sensitivity to conditional context rather than reliable answer-aligned optimization directions.

\paragraph{Overlap in Counterfactual Optimization Signals.}
Table~\ref{tab:alternative_cpc_statistics} reports the trajectory-level combined signal magnitude $\mathcal{M}$, disagreement magnitude $\mathcal{D}$, and Counterfactual Perturbation Consistency (CPC) under the three alternative prompt constructions.

\begin{table}[t]
\centering
\caption{
Trajectory-level optimization-signal statistics under alternative counterfactual privileged prompts, computed over the same 400 fixed on-policy trajectories.
}
\label{tab:alternative_cpc_statistics}
\scriptsize
\setlength{\tabcolsep}{3.2pt}
\renewcommand{\arraystretch}{1.04}
\begin{tabular*}{\columnwidth}{
@{\extracolsep{\fill}}
llrrrr
@{}
}
\toprule
\textbf{Construction}
& \textbf{Metric}
& \textbf{Mean}
& \textbf{Median}
& \textbf{Minimum}
& \textbf{Maximum}
\\
\midrule

Answer-level
& $\mathcal{M}$ & 0.059 & 0.055 & 0.024 & 0.157 \\
& $\mathcal{D}$ & 0.021 & 0.019 & 0.009 & 0.060 \\
& CPC           & 0.636 & 0.636 & 0.489 & 0.817 \\
\addlinespace[2pt]

Mild outcome-only
& $\mathcal{M}$ & 0.098 & 0.095 & 0.050 & 0.214 \\
& $\mathcal{D}$ & 0.020 & 0.019 & 0.011 & 0.050 \\
& CPC           & 0.789 & 0.794 & 0.691 & 0.856 \\
\addlinespace[2pt]

Reference solution
& $\mathcal{M}$ & 0.138 & 0.131 & 0.050 & 0.495 \\
& $\mathcal{D}$ & 0.064 & 0.061 & 0.018 & 0.222 \\
& CPC           & 0.533 & 0.535 & 0.308 & 0.719 \\
\bottomrule
\end{tabular*}
\end{table}

All three constructions produce substantial same-coordinate, same-direction overlap between the optimization signals induced by their positive and negative conditions. Answer-level feedback yields a mean CPC of $0.636$, indicating that, on average across trajectories, nearly two thirds of the combined full-vocabulary signal magnitude lies on the same vocabulary coordinates and points in the same direction, despite reversing the asserted correctness judgment assigned to the same ground-truth answer. The overlap is considerably stronger under mild outcome-only feedback, for which the mean and median CPC reach $0.789$ and $0.794$, respectively, while even the minimum trajectory-level CPC remains $0.691$. Removing the polarized praise and criticism from the main-text prompts therefore increases, rather than decreases, the same-coordinate, same-direction agreement between the two opposing privileged signals.

Reference-solution conditioning induces the largest absolute perturbations, with mean $\mathcal{M}=0.138$ and $\mathcal{D}=0.064$, consistent with the greater content-level difference introduced by supplying complete and semantically distinct reasoning processes. Nevertheless, its mean and median CPC remain above $0.53$, showing that more than half of the combined signal magnitude still overlaps on average across trajectories. Thus, even rich correct and incorrect reference solutions fail to produce predominantly opposing optimization signals. Together with the full-vocabulary directional statistics in Figure~\ref{fig:alternative_full_vocab_shift_composition}, these results demonstrate that substantial counterfactual overlap persists across answer-level, outcome-level, and full-reasoning privileged information.

\begin{table*}[!t]
\caption{
Top-50 most frequent sampled tokens with large positive, large negative, or near-zero likelihood shifts under the answer-level counterfactual privileged prompts, computed over the same 400 fixed on-policy trajectories. Pure-whitespace tokens are excluded. Numbers in parentheses denote occurrence counts.
}
\label{tab:answer_level_top50_shift_tokens}
\centering
\scriptsize
\setlength{\tabcolsep}{1.7pt}
\renewcommand{\arraystretch}{0.82}

\newcommand{\tokans}[2]{\texttt{#1}\,(#2)}

\begin{tabular*}{\textwidth}{
@{\extracolsep{\fill}}
r
llllll
@{}
}
\toprule
&
\multicolumn{2}{c}{\textbf{Large Positive Shifts}}
&
\multicolumn{2}{c}{\textbf{Large Negative Shifts}}
&
\multicolumn{2}{c}{\textbf{Near-Zero Shifts}}
\\
\cmidrule(lr){2-3}
\cmidrule(lr){4-5}
\cmidrule(lr){6-7}
\textbf{Rank}
&
$\boldsymbol{z^{\mathrm{pos}}>0.1}$
&
$\boldsymbol{z^{\mathrm{neg}}>0.1}$
&
$\boldsymbol{z^{\mathrm{pos}}<-0.1}$
&
$\boldsymbol{z^{\mathrm{neg}}<-0.1}$
&
$\boldsymbol{|z^{\mathrm{pos}}|<0.01}$
&
$\boldsymbol{|z^{\mathrm{neg}}|<0.01}$
\\
\midrule

1
& \tokans{The}{6,787}
& \tokans{The}{17,515}
& \tokans{.}{7,911}
& \tokans{.}{9,833}
& \tokans{,}{173,455}
& \tokans{,}{152,663}
\\

2
& \tokans{.}{5,948}
& \tokans{.}{16,347}
& \tokans{The}{5,209}
& \tokans{The}{6,812}
& \tokans{1}{152,346}
& \tokans{1}{149,569}
\\

3
& \tokans{But}{3,213}
& \tokans{,}{9,952}
& \tokans{But}{3,583}
& \tokans{Is}{6,474}
& \tokans{2}{145,236}
& \tokans{2}{142,884}
\\

4
& \tokans{,}{3,075}
& \tokans{So}{8,135}
& \tokans{Let}{3,554}
& \tokans{,}{5,075}
& \tokans{0}{110,422}
& \tokans{0}{109,404}
\\

5
& \tokans{So}{2,436}
& \tokans{But}{7,495}
& \tokans{,}{3,506}
& \tokans{So}{4,569}
& \tokans{.}{95,911}
& \tokans{=}{85,863}
\\

6
& \tokans{Let}{1,899}
& \tokans{Let}{5,668}
& \tokans{So}{3,274}
& \tokans{Therefore}{4,433}
& \tokans{=}{91,936}
& \tokans{3}{80,888}
\\

7
& \tokans{Is}{1,760}
& \tokans{For}{4,683}
& \tokans{Is}{3,043}
& \tokans{Wait}{3,746}
& \tokans{3}{82,233}
& \tokans{.}{66,598}
\\

8
& \tokans{For}{1,318}
& \tokans{'s}{3,854}
& \tokans{Wait}{2,822}
& \tokans{But}{3,742}
& \tokans{The}{71,076}
& \tokans{5}{60,321}
\\

9
& \tokans{And}{1,251}
& \tokans{This}{3,812}
& \tokans{Therefore}{2,372}
& \tokans{Let}{2,673}
& \tokans{5}{61,061}
& \tokans{+}{56,526}
\\

10
& \tokans{This}{1,230}
& \tokans{And}{3,788}
& \tokans{I}{1,853}
& \tokans{:}{2,580}
& \tokans{Is}{57,928}
& \tokans{4}{55,020}
\\

11
& \tokans{If}{1,203}
& \tokans{If}{3,648}
& \tokans{If}{1,767}
& \tokans{?}{2,404}
& \tokans{+}{57,542}
& \tokans{-}{52,068}
\\

12
& \tokans{Therefore}{1,166}
& \tokans{(}{3,356}
& \tokans{Maybe}{1,535}
& \tokans{That}{2,283}
& \tokans{(}{56,672}
& \tokans{(}{50,766}
\\

13
& \tokans{A}{1,143}
& \tokans{A}{3,250}
& \tokans{A}{1,492}
& \tokans{If}{2,030}
& \tokans{4}{55,788}
& \tokans{The}{48,552}
\\

14
& \tokans{(}{1,136}
& \tokans{There}{3,228}
& \tokans{That}{1,442}
& \tokans{Maybe}{1,953}
& \tokans{-}{53,748}
& \tokans{Is}{48,275}
\\

15
& \tokans{That}{1,128}
& \tokans{It}{2,848}
& \tokans{We}{1,286}
& \tokans{We}{1,940}
& \tokans{6}{45,550}
& \tokans{6}{44,837}
\\

16
& \tokans{=}{1,065}
& \tokans{:}{2,788}
& \tokans{For}{1,266}
& \tokans{Then}{1,889}
& \tokans{)}{43,340}
& \tokans{)}{40,164}
\\

17
& \tokans{:}{1,054}
& \tokans{=}{2,706}
& \tokans{?}{1,226}
& \tokans{Alternatively}{1,868}
& \tokans{/}{37,521}
& \tokans{/}{35,891}
\\

18
& \tokans{Wait}{1,014}
& \tokans{That}{2,528}
& \tokans{(}{1,196}
& \tokans{I}{1,636}
& \tokans{A}{36,175}
& \tokans{A}{30,818}
\\

19
& \tokans{There}{864}
& \tokans{Is}{2,336}
& \tokans{In}{1,152}
& \tokans{A}{1,621}
& \tokans{Of}{33,095}
& \tokans{<U+FFFD>}{29,945}
\\

20
& \tokans{In}{826}
& \tokans{In}{2,277}
& \tokans{:}{1,121}
& \tokans{(}{1,596}
& \tokans{<U+FFFD>}{31,079}
& \tokans{Of}{29,172}
\\

21
& \tokans{Maybe}{793}
& \tokans{No}{1,994}
& \tokans{And}{1,034}
& \tokans{Me}{1,500}
& \tokans{:}{29,353}
& \tokans{8}{26,086}
\\

22
& \tokans{Check}{790}
& \tokans{Maybe}{1,974}
& \tokans{Alternatively}{1,007}
& \tokans{In}{1,465}
& \tokans{8}{26,414}
& \tokans{9}{25,625}
\\

23
& \tokans{We}{775}
& \tokans{We}{1,965}
& \tokans{Since}{966}
& \tokans{Since}{1,337}
& \tokans{9}{26,003}
& \tokans{7}{24,972}
\\

24
& \tokans{'s}{725}
& \tokans{Since}{1,952}
& \tokans{There}{951}
& \tokans{And}{1,275}
& \tokans{To}{25,867}
& \tokans{:}{24,087}
\\

25
& \tokans{1}{720}
& \tokans{Therefore}{1,947}
& \tokans{Then}{894}
& \tokans{Of}{1,272}
& \tokans{7}{25,307}
& \tokans{To}{23,460}
\\

26
& \tokans{Since}{716}
& \tokans{Wait}{1,943}
& \tokans{=}{842}
& \tokans{=}{1,253}
& \tokans{And}{21,358}
& \tokans{N}{16,949}
\\

27
& \tokans{Then}{703}
& \tokans{Each}{1,806}
& \tokans{Of}{835}
& \tokans{Which}{1,167}
& \tokans{*}{18,851}
& \tokans{\$}{16,930}
\\

28
& \tokans{It}{632}
& \tokans{I}{1,767}
& \tokans{1}{834}
& \tokans{For}{1,128}
& \tokans{N}{18,502}
& \tokans{B}{16,918}
\\

29
& \tokans{?}{605}
& \tokans{Then}{1,654}
& \tokans{Problem}{782}
& \tokans{However}{1,093}
& \tokans{\$}{18,089}
& \tokans{*}{16,731}
\\

30
& \tokans{Of}{575}
& \tokans{Would}{1,612}
& \tokans{This}{781}
& \tokans{Check}{1,078}
& \tokans{B}{17,688}
& \tokans{And}{16,280}
\\

31
& \tokans{2}{558}
& \tokans{To}{1,593}
& \tokans{Which}{769}
& \tokans{With}{1,071}
& \tokans{X}{17,427}
& \tokans{\textbackslash\textbackslash}{16,266}
\\

32
& \tokans{Each}{524}
& \tokans{1}{1,397}
& \tokans{With}{748}
& \tokans{1}{1,023}
& \tokans{\textbackslash\textbackslash}{16,406}
& \tokans{X}{16,134}
\\

33
& \tokans{I}{513}
& \tokans{Of}{1,382}
& \tokans{Think}{717}
& \tokans{).}{972}
& \tokans{So}{14,918}
& \tokans{\textsuperscript{2}}{14,350}
\\

34
& \tokans{Which}{508}
& \tokans{Which}{1,375}
& \tokans{2}{640}
& \tokans{Compute}{953}
& \tokans{\textsuperscript{2}}{14,554}
& \tokans{K}{13,284}
\\

35
& \tokans{Problem}{502}
& \tokans{Are}{1,273}
& \tokans{To}{635}
& \tokans{Not}{892}
& \tokans{K}{14,340}
& \tokans{\}}{12,514}
\\

36
& \tokans{To}{502}
& \tokans{?}{1,266}
& \tokans{However}{607}
& \tokans{Think}{887}
& \tokans{Are}{13,934}
& \tokans{Y}{11,698}
\\

37
& \tokans{Total}{500}
& \tokans{)}{1,251}
& \tokans{Number}{546}
& \tokans{Answer}{833}
& \tokans{That}{12,943}
& \tokans{C}{11,382}
\\

38
& \tokans{With}{499}
& \tokans{From}{1,244}
& \tokans{First}{520}
& \tokans{Can}{822}
& \tokans{\}}{12,903}
& \tokans{Are}{11,234}
\\

39
& \tokans{No}{446}
& \tokans{Total}{1,183}
& \tokans{Each}{518}
& \tokans{Problem}{806}
& \tokans{Me}{12,512}
& \tokans{Me}{11,016}
\\

40
& \tokans{)}{437}
& \tokans{Where}{1,140}
& \tokans{Can}{514}
& \tokans{First}{788}
& \tokans{Y}{12,206}
& \tokans{\ensuremath{\sqrt{\phantom{0}}}}{10,974}
\\

41
& \tokans{Would}{428}
& \tokans{2}{1,113}
& \tokans{N}{511}
& \tokans{Number}{780}
& \tokans{C}{12,196}
& \tokans{\{}{10,535}
\\

42
& \tokans{-}{426}
& \tokans{With}{1,068}
& \tokans{Answer}{499}
& \tokans{2}{759}
& \tokans{But}{11,312}
& \tokans{Be}{10,472}
\\

43
& \tokans{Are}{426}
& \tokans{-}{1,065}
& \tokans{It}{497}
& \tokans{To}{715}
& \tokans{Be}{11,252}
& \tokans{\textasciicircum}{9,655}
\\

44
& \tokans{N}{416}
& \tokans{When}{1,003}
& \tokans{3}{496}
& \tokans{-}{649}
& \tokans{\ensuremath{\sqrt{\phantom{0}}}}{11,061}
& \tokans{That}{9,585}
\\

45
& \tokans{From}{409}
& \tokans{3}{964}
& \tokans{Check}{483}
& \tokans{Have}{637}
& \tokans{\{}{10,636}
& \tokans{M}{8,975}
\\

46
& \tokans{3}{407}
& \tokans{N}{951}
& \tokans{'s}{482}
& \tokans{)}{603}
& \tokans{\textasciicircum}{10,091}
& \tokans{R}{8,527}
\\

47
& \tokans{However}{404}
& \tokans{),}{927}
& \tokans{Not}{462}
& \tokans{Need}{601}
& \tokans{In}{9,857}
& \tokans{P}{7,515}
\\

48
& \tokans{Can}{359}
& \tokans{At}{915}
& \tokans{-}{436}
& \tokans{Are}{597}
& \tokans{M}{9,780}
& \tokans{T}{7,461}
\\

49
& \tokans{Alternatively}{357}
& \tokans{X}{915}
& \tokans{Seems}{432}
& \tokans{N}{588}
& \tokans{Then}{9,245}
& \tokans{So}{7,095}
\\

50
& \tokans{Number}{354}
& \tokans{*}{867}
& \tokans{Are}{397}
& \tokans{/}{567}
& \tokans{R}{9,177}
& \tokans{In}{7,005}
\\

\bottomrule
\end{tabular*}
\end{table*}

\begin{table*}[!t]
\caption{
Top-50 most frequent sampled tokens with large positive, large negative, or near-zero likelihood shifts under the mild outcome-only counterfactual privileged prompts, computed over the same 400 fixed on-policy trajectories. Pure-whitespace tokens are excluded. Numbers in parentheses denote occurrence counts.
}
\label{tab:outcome_only_top50_shift_tokens}
\centering
\scriptsize
\setlength{\tabcolsep}{1.7pt}
\renewcommand{\arraystretch}{0.82}

\newcommand{\tokout}[2]{\texttt{#1}\,(#2)}

\begin{tabular*}{\textwidth}{
@{\extracolsep{\fill}}
r
llllll
@{}
}
\toprule
&
\multicolumn{2}{c}{\textbf{Large Positive Shifts}}
&
\multicolumn{2}{c}{\textbf{Large Negative Shifts}}
&
\multicolumn{2}{c}{\textbf{Near-Zero Shifts}}
\\
\cmidrule(lr){2-3}
\cmidrule(lr){4-5}
\cmidrule(lr){6-7}
\textbf{Rank}
&
$\boldsymbol{z^{\mathrm{pos}}>0.1}$
&
$\boldsymbol{z^{\mathrm{neg}}>0.1}$
&
$\boldsymbol{z^{\mathrm{pos}}<-0.1}$
&
$\boldsymbol{z^{\mathrm{neg}}<-0.1}$
&
$\boldsymbol{|z^{\mathrm{pos}}|<0.01}$
&
$\boldsymbol{|z^{\mathrm{neg}}|<0.01}$
\\
\midrule

1
& \tokout{The}{17,515}
& \tokout{The}{17,275}
& \tokout{.}{9,833}
& \tokout{.}{10,493}
& \tokout{,}{152,663}
& \tokout{,}{151,870}
\\

2
& \tokout{.}{16,347}
& \tokout{.}{16,945}
& \tokout{The}{6,812}
& \tokout{The}{7,564}
& \tokout{1}{149,569}
& \tokout{1}{149,501}
\\

3
& \tokout{,}{9,952}
& \tokout{,}{11,044}
& \tokout{Is}{6,474}
& \tokout{Is}{6,798}
& \tokout{2}{142,884}
& \tokout{2}{142,878}
\\

4
& \tokout{So}{8,135}
& \tokout{But}{8,210}
& \tokout{,}{5,075}
& \tokout{,}{5,013}
& \tokout{0}{109,404}
& \tokout{0}{109,426}
\\

5
& \tokout{But}{7,495}
& \tokout{So}{8,105}
& \tokout{So}{4,569}
& \tokout{So}{4,781}
& \tokout{=}{85,863}
& \tokout{=}{85,763}
\\

6
& \tokout{Let}{5,668}
& \tokout{Let}{5,612}
& \tokout{Therefore}{4,433}
& \tokout{Therefore}{4,680}
& \tokout{3}{80,888}
& \tokout{3}{80,861}
\\

7
& \tokout{For}{4,683}
& \tokout{For}{5,084}
& \tokout{Wait}{3,746}
& \tokout{Wait}{3,754}
& \tokout{.}{66,598}
& \tokout{.}{65,525}
\\

8
& \tokout{'s}{3,854}
& \tokout{This}{4,223}
& \tokout{But}{3,742}
& \tokout{Let}{3,309}
& \tokout{5}{60,321}
& \tokout{5}{60,274}
\\

9
& \tokout{This}{3,812}
& \tokout{'s}{3,951}
& \tokout{Let}{2,673}
& \tokout{But}{3,259}
& \tokout{+}{56,526}
& \tokout{+}{56,433}
\\

10
& \tokout{And}{3,788}
& \tokout{If}{3,813}
& \tokout{:}{2,580}
& \tokout{:}{2,569}
& \tokout{4}{55,020}
& \tokout{4}{55,087}
\\

11
& \tokout{If}{3,648}
& \tokout{And}{3,687}
& \tokout{?}{2,404}
& \tokout{?}{2,421}
& \tokout{-}{52,068}
& \tokout{-}{51,971}
\\

12
& \tokout{(}{3,356}
& \tokout{There}{3,403}
& \tokout{That}{2,283}
& \tokout{That}{2,406}
& \tokout{(}{50,766}
& \tokout{(}{50,651}
\\

13
& \tokout{A}{3,250}
& \tokout{(}{3,401}
& \tokout{If}{2,030}
& \tokout{We}{2,170}
& \tokout{The}{48,552}
& \tokout{The}{48,383}
\\

14
& \tokout{There}{3,228}
& \tokout{A}{3,388}
& \tokout{Maybe}{1,953}
& \tokout{If}{2,155}
& \tokout{Is}{48,275}
& \tokout{Is}{47,619}
\\

15
& \tokout{It}{2,848}
& \tokout{:}{3,134}
& \tokout{We}{1,940}
& \tokout{Maybe}{2,037}
& \tokout{6}{44,837}
& \tokout{6}{44,876}
\\

16
& \tokout{:}{2,788}
& \tokout{It}{2,955}
& \tokout{Then}{1,889}
& \tokout{Alternatively}{1,951}
& \tokout{)}{40,164}
& \tokout{)}{40,095}
\\

17
& \tokout{=}{2,706}
& \tokout{=}{2,833}
& \tokout{Alternatively}{1,868}
& \tokout{Then}{1,824}
& \tokout{/}{35,891}
& \tokout{/}{35,838}
\\

18
& \tokout{That}{2,528}
& \tokout{That}{2,597}
& \tokout{I}{1,636}
& \tokout{A}{1,737}
& \tokout{A}{30,818}
& \tokout{A}{30,707}
\\

19
& \tokout{Is}{2,336}
& \tokout{Is}{2,542}
& \tokout{A}{1,621}
& \tokout{Me}{1,697}
& \tokout{<U+FFFD>}{29,945}
& \tokout{<U+FFFD>}{29,869}
\\

20
& \tokout{In}{2,277}
& \tokout{In}{2,462}
& \tokout{(}{1,596}
& \tokout{I}{1,667}
& \tokout{Of}{29,172}
& \tokout{Of}{29,123}
\\

21
& \tokout{No}{1,994}
& \tokout{No}{2,133}
& \tokout{Me}{1,500}
& \tokout{(}{1,618}
& \tokout{8}{26,086}
& \tokout{8}{26,094}
\\

22
& \tokout{Maybe}{1,974}
& \tokout{Wait}{2,126}
& \tokout{In}{1,465}
& \tokout{Since}{1,551}
& \tokout{9}{25,625}
& \tokout{9}{25,608}
\\

23
& \tokout{We}{1,965}
& \tokout{Maybe}{2,013}
& \tokout{Since}{1,337}
& \tokout{In}{1,489}
& \tokout{7}{24,972}
& \tokout{7}{24,997}
\\

24
& \tokout{Since}{1,952}
& \tokout{Since}{1,967}
& \tokout{And}{1,275}
& \tokout{And}{1,466}
& \tokout{:}{24,087}
& \tokout{:}{23,813}
\\

25
& \tokout{Therefore}{1,947}
& \tokout{We}{1,886}
& \tokout{Of}{1,272}
& \tokout{Of}{1,387}
& \tokout{To}{23,460}
& \tokout{To}{23,327}
\\

26
& \tokout{Wait}{1,943}
& \tokout{Each}{1,873}
& \tokout{=}{1,253}
& \tokout{=}{1,313}
& \tokout{N}{16,949}
& \tokout{N}{16,934}
\\

27
& \tokout{Each}{1,806}
& \tokout{Then}{1,855}
& \tokout{Which}{1,167}
& \tokout{Which}{1,185}
& \tokout{\$}{16,930}
& \tokout{\$}{16,914}
\\

28
& \tokout{I}{1,767}
& \tokout{I}{1,841}
& \tokout{For}{1,128}
& \tokout{With}{1,185}
& \tokout{B}{16,918}
& \tokout{B}{16,891}
\\

29
& \tokout{Then}{1,654}
& \tokout{Therefore}{1,791}
& \tokout{However}{1,093}
& \tokout{1}{1,057}
& \tokout{*}{16,731}
& \tokout{*}{16,685}
\\

30
& \tokout{Would}{1,612}
& \tokout{Would}{1,736}
& \tokout{Check}{1,078}
& \tokout{For}{1,054}
& \tokout{And}{16,280}
& \tokout{\textbackslash\textbackslash}{16,310}
\\

31
& \tokout{To}{1,593}
& \tokout{To}{1,533}
& \tokout{With}{1,071}
& \tokout{).}{1,023}
& \tokout{\textbackslash\textbackslash}{16,266}
& \tokout{And}{16,187}
\\

32
& \tokout{1}{1,397}
& \tokout{Which}{1,455}
& \tokout{1}{1,023}
& \tokout{Compute}{1,021}
& \tokout{X}{16,134}
& \tokout{X}{16,138}
\\

33
& \tokout{Of}{1,382}
& \tokout{1}{1,426}
& \tokout{).}{972}
& \tokout{Check}{995}
& \tokout{\textsuperscript{2}}{14,350}
& \tokout{\textsuperscript{2}}{14,278}
\\

34
& \tokout{Which}{1,375}
& \tokout{Of}{1,402}
& \tokout{Compute}{953}
& \tokout{However}{975}
& \tokout{K}{13,284}
& \tokout{K}{13,271}
\\

35
& \tokout{Are}{1,273}
& \tokout{?}{1,394}
& \tokout{Not}{892}
& \tokout{Not}{970}
& \tokout{\}}{12,514}
& \tokout{\}}{12,540}
\\

36
& \tokout{?}{1,266}
& \tokout{Are}{1,380}
& \tokout{Think}{887}
& \tokout{Think}{953}
& \tokout{Y}{11,698}
& \tokout{Y}{11,668}
\\

37
& \tokout{)}{1,251}
& \tokout{)}{1,313}
& \tokout{Answer}{833}
& \tokout{Can}{912}
& \tokout{C}{11,382}
& \tokout{C}{11,353}
\\

38
& \tokout{From}{1,244}
& \tokout{From}{1,245}
& \tokout{Can}{822}
& \tokout{Number}{836}
& \tokout{Are}{11,234}
& \tokout{Are}{11,127}
\\

39
& \tokout{Total}{1,183}
& \tokout{Total}{1,215}
& \tokout{Problem}{806}
& \tokout{First}{832}
& \tokout{Me}{11,016}
& \tokout{\ensuremath{\sqrt{\phantom{0}}}}{10,974}
\\

40
& \tokout{Where}{1,140}
& \tokout{Where}{1,185}
& \tokout{First}{788}
& \tokout{To}{830}
& \tokout{\ensuremath{\sqrt{\phantom{0}}}}{10,974}
& \tokout{Me}{10,807}
\\

41
& \tokout{2}{1,113}
& \tokout{-}{1,133}
& \tokout{Number}{780}
& \tokout{Problem}{811}
& \tokout{\{}{10,535}
& \tokout{\{}{10,544}
\\

42
& \tokout{With}{1,068}
& \tokout{With}{1,088}
& \tokout{2}{759}
& \tokout{2}{790}
& \tokout{Be}{10,472}
& \tokout{Be}{10,423}
\\

43
& \tokout{-}{1,065}
& \tokout{2}{1,079}
& \tokout{To}{715}
& \tokout{Answer}{779}
& \tokout{\textasciicircum}{9,655}
& \tokout{\textasciicircum}{9,652}
\\

44
& \tokout{When}{1,003}
& \tokout{At}{1,013}
& \tokout{-}{649}
& \tokout{-}{708}
& \tokout{That}{9,585}
& \tokout{That}{9,415}
\\

45
& \tokout{3}{964}
& \tokout{),}{997}
& \tokout{Have}{637}
& \tokout{Have}{675}
& \tokout{M}{8,975}
& \tokout{M}{8,948}
\\

46
& \tokout{N}{951}
& \tokout{When}{992}
& \tokout{)}{603}
& \tokout{Need}{652}
& \tokout{R}{8,527}
& \tokout{R}{8,541}
\\

47
& \tokout{),}{927}
& \tokout{N}{966}
& \tokout{Need}{601}
& \tokout{/}{640}
& \tokout{P}{7,515}
& \tokout{P}{7,530}
\\

48
& \tokout{At}{915}
& \tokout{3}{953}
& \tokout{Are}{597}
& \tokout{)}{627}
& \tokout{T}{7,461}
& \tokout{T}{7,418}
\\

49
& \tokout{X}{915}
& \tokout{*}{909}
& \tokout{N}{588}
& \tokout{Again}{619}
& \tokout{So}{7,095}
& \tokout{So}{7,013}
\\

50
& \tokout{*}{867}
& \tokout{X}{907}
& \tokout{/}{567}
& \tokout{N}{604}
& \tokout{In}{7,005}
& \tokout{In}{6,888}
\\

\bottomrule
\end{tabular*}
\end{table*}

\begin{table*}[!t]
\caption{
Top-50 most frequent sampled tokens with large positive, large negative, or near-zero likelihood shifts under the correct/incorrect reference-solution counterfactual privileged prompts, computed over the same 400 fixed on-policy trajectories. Pure-whitespace tokens are excluded. Numbers in parentheses denote occurrence counts.
}
\label{tab:reference_solution_top50_shift_tokens}
\centering
\scriptsize
\setlength{\tabcolsep}{1.7pt}
\renewcommand{\arraystretch}{0.82}

\newcommand{\toksol}[2]{\texttt{#1}\,(#2)}

\begin{tabular*}{\textwidth}{
@{\extracolsep{\fill}}
r
llllll
@{}
}
\toprule
&
\multicolumn{2}{c}{\textbf{Large Positive Shifts}}
&
\multicolumn{2}{c}{\textbf{Large Negative Shifts}}
&
\multicolumn{2}{c}{\textbf{Near-Zero Shifts}}
\\
\cmidrule(lr){2-3}
\cmidrule(lr){4-5}
\cmidrule(lr){6-7}
\textbf{Rank}
&
$\boldsymbol{z^{\mathrm{pos}}>0.1}$
&
$\boldsymbol{z^{\mathrm{neg}}>0.1}$
&
$\boldsymbol{z^{\mathrm{pos}}<-0.1}$
&
$\boldsymbol{z^{\mathrm{neg}}<-0.1}$
&
$\boldsymbol{|z^{\mathrm{pos}}|<0.01}$
&
$\boldsymbol{|z^{\mathrm{neg}}|<0.01}$
\\
\midrule

1
& \toksol{The}{15,368}
& \toksol{The}{14,695}
& \toksol{.}{13,990}
& \toksol{.}{12,874}
& \toksol{,}{161,071}
& \toksol{,}{162,083}
\\

2
& \toksol{.}{10,340}
& \toksol{.}{10,384}
& \toksol{The}{10,632}
& \toksol{The}{10,032}
& \toksol{1}{149,683}
& \toksol{1}{149,980}
\\

3
& \toksol{,}{5,996}
& \toksol{,}{5,600}
& \toksol{,}{7,578}
& \toksol{,}{7,280}
& \toksol{2}{143,095}
& \toksol{2}{143,346}
\\

4
& \toksol{So}{5,033}
& \toksol{But}{5,266}
& \toksol{Let}{6,926}
& \toksol{Let}{6,657}
& \toksol{0}{109,371}
& \toksol{0}{109,554}
\\

5
& \toksol{But}{4,820}
& \toksol{So}{4,978}
& \toksol{But}{6,652}
& \toksol{Is}{6,395}
& \toksol{=}{87,856}
& \toksol{=}{88,269}
\\

6
& \toksol{Is}{3,462}
& \toksol{Is}{3,078}
& \toksol{Is}{6,391}
& \toksol{Therefore}{6,048}
& \toksol{3}{80,903}
& \toksol{3}{81,242}
\\

7
& \toksol{Let}{2,698}
& \toksol{Let}{2,667}
& \toksol{So}{6,163}
& \toksol{So}{5,835}
& \toksol{.}{79,330}
& \toksol{.}{80,155}
\\

8
& \toksol{For}{2,598}
& \toksol{A}{2,588}
& \toksol{Therefore}{5,776}
& \toksol{But}{5,551}
& \toksol{5}{60,337}
& \toksol{5}{60,413}
\\

9
& \toksol{A}{2,578}
& \toksol{And}{2,378}
& \toksol{Wait}{4,978}
& \toksol{If}{4,323}
& \toksol{+}{56,356}
& \toksol{+}{56,475}
\\

10
& \toksol{And}{2,496}
& \toksol{This}{2,315}
& \toksol{If}{4,527}
& \toksol{Wait}{4,244}
& \toksol{4}{55,048}
& \toksol{The}{55,625}
\\

11
& \toksol{(}{2,355}
& \toksol{For}{2,314}
& \toksol{Maybe}{4,283}
& \toksol{Maybe}{3,928}
& \toksol{The}{54,457}
& \toksol{4}{55,159}
\\

12
& \toksol{This}{2,227}
& \toksol{(}{2,296}
& \toksol{A}{3,471}
& \toksol{We}{3,113}
& \toksol{(}{52,462}
& \toksol{(}{53,113}
\\

13
& \toksol{=}{2,226}
& \toksol{=}{2,181}
& \toksol{We}{3,264}
& \toksol{A}{3,111}
& \toksol{-}{52,206}
& \toksol{-}{52,397}
\\

14
& \toksol{That}{1,915}
& \toksol{Wait}{2,156}
& \toksol{?}{3,226}
& \toksol{?}{3,085}
& \toksol{Is}{49,555}
& \toksol{Is}{49,832}
\\

15
& \toksol{In}{1,862}
& \toksol{That}{1,970}
& \toksol{That}{3,128}
& \toksol{That}{3,029}
& \toksol{6}{44,877}
& \toksol{6}{44,962}
\\

16
& \toksol{'s}{1,757}
& \toksol{:}{1,711}
& \toksol{:}{3,126}
& \toksol{I}{3,019}
& \toksol{)}{41,264}
& \toksol{)}{41,393}
\\

17
& \toksol{:}{1,744}
& \toksol{'s}{1,655}
& \toksol{I}{3,043}
& \toksol{:}{2,967}
& \toksol{/}{36,187}
& \toksol{/}{36,298}
\\

18
& \toksol{Wait}{1,733}
& \toksol{In}{1,655}
& \toksol{(}{2,836}
& \toksol{For}{2,677}
& \toksol{A}{31,886}
& \toksol{A}{32,336}
\\

19
& \toksol{If}{1,572}
& \toksol{If}{1,514}
& \toksol{Since}{2,602}
& \toksol{Since}{2,570}
& \toksol{<U+FFFD>}{30,037}
& \toksol{<U+FFFD>}{30,173}
\\

20
& \toksol{We}{1,339}
& \toksol{Check}{1,425}
& \toksol{For}{2,556}
& \toksol{Alternatively}{2,543}
& \toksol{Of}{29,785}
& \toksol{Of}{30,001}
\\

21
& \toksol{1}{1,332}
& \toksol{Of}{1,284}
& \toksol{Alternatively}{2,554}
& \toksol{(}{2,445}
& \toksol{8}{26,077}
& \toksol{8}{26,138}
\\

22
& \toksol{Of}{1,292}
& \toksol{We}{1,252}
& \toksol{There}{2,384}
& \toksol{There}{2,312}
& \toksol{:}{25,666}
& \toksol{:}{25,831}
\\

23
& \toksol{Check}{1,238}
& \toksol{It}{1,229}
& \toksol{Of}{2,283}
& \toksol{Then}{2,225}
& \toksol{9}{25,584}
& \toksol{9}{25,681}
\\

24
& \toksol{It}{1,232}
& \toksol{1}{1,221}
& \toksol{And}{2,267}
& \toksol{In}{2,189}
& \toksol{7}{24,978}
& \toksol{7}{25,020}
\\

25
& \toksol{Since}{1,208}
& \toksol{Then}{1,099}
& \toksol{Then}{2,252}
& \toksol{And}{2,169}
& \toksol{To}{24,138}
& \toksol{To}{24,198}
\\

26
& \toksol{Then}{1,148}
& \toksol{Since}{1,090}
& \toksol{In}{2,239}
& \toksol{Of}{2,167}
& \toksol{And}{18,026}
& \toksol{And}{18,307}
\\

27
& \toksol{To}{1,140}
& \toksol{To}{1,027}
& \toksol{1}{2,049}
& \toksol{1}{1,922}
& \toksol{*}{17,534}
& \toksol{*}{17,513}
\\

28
& \toksol{2}{1,099}
& \toksol{2}{995}
& \toksol{Problem}{1,866}
& \toksol{Problem}{1,874}
& \toksol{N}{17,255}
& \toksol{N}{17,418}
\\

29
& \toksol{There}{1,021}
& \toksol{There}{990}
& \toksol{=}{1,857}
& \toksol{Think}{1,654}
& \toksol{B}{17,083}
& \toksol{B}{17,236}
\\

30
& \toksol{Are}{1,012}
& \toksol{Each}{982}
& \toksol{Think}{1,769}
& \toksol{With}{1,642}
& \toksol{X}{16,406}
& \toksol{X}{16,618}
\\

31
& \toksol{I}{999}
& \toksol{Are}{956}
& \toksol{With}{1,701}
& \toksol{Which}{1,578}
& \toksol{\$}{16,068}
& \toksol{\$}{16,403}
\\

32
& \toksol{Therefore}{999}
& \toksol{Maybe}{928}
& \toksol{Which}{1,656}
& \toksol{=}{1,576}
& \toksol{\textbackslash\textbackslash}{15,598}
& \toksol{\textbackslash\textbackslash}{15,780}
\\

33
& \toksol{Each}{911}
& \toksol{I}{922}
& \toksol{2}{1,575}
& \toksol{2}{1,457}
& \toksol{\textsuperscript{2}}{14,227}
& \toksol{\textsuperscript{2}}{14,262}
\\

34
& \toksol{)}{904}
& \toksol{)}{907}
& \toksol{Number}{1,414}
& \toksol{Me}{1,320}
& \toksol{K}{13,440}
& \toksol{K}{13,706}
\\

35
& \toksol{With}{862}
& \toksol{Therefore}{827}
& \toksol{Each}{1,412}
& \toksol{Number}{1,304}
& \toksol{\}}{12,462}
& \toksol{\}}{12,544}
\\

36
& \toksol{From}{846}
& \toksol{With}{803}
& \toksol{First}{1,399}
& \toksol{First}{1,270}
& \toksol{Are}{12,064}
& \toksol{Are}{12,175}
\\

37
& \toksol{Maybe}{817}
& \toksol{From}{799}
& \toksol{This}{1,365}
& \toksol{To}{1,270}
& \toksol{Y}{11,799}
& \toksol{Y}{11,918}
\\

38
& \toksol{Which}{815}
& \toksol{No}{759}
& \toksol{Me}{1,270}
& \toksol{Each}{1,257}
& \toksol{C}{11,503}
& \toksol{C}{11,614}
\\

39
& \toksol{\$}{782}
& \toksol{Which}{749}
& \toksol{However}{1,249}
& \toksol{Can}{1,196}
& \toksol{\ensuremath{\sqrt{\phantom{0}}}}{10,948}
& \toksol{\ensuremath{\sqrt{\phantom{0}}}}{10,979}
\\

40
& \toksol{No}{757}
& \toksol{-}{726}
& \toksol{To}{1,231}
& \toksol{However}{1,153}
& \toksol{Be}{10,494}
& \toksol{Be}{10,543}
\\

41
& \toksol{-}{737}
& \toksol{3}{708}
& \toksol{N}{1,224}
& \toksol{).}{1,144}
& \toksol{\{}{10,487}
& \toksol{\{}{10,537}
\\

42
& \toksol{3}{688}
& \toksol{?}{707}
& \toksol{Can}{1,160}
& \toksol{This}{1,117}
& \toksol{That}{10,279}
& \toksol{That}{10,402}
\\

43
& \toksol{Can}{681}
& \toksol{\$}{669}
& \toksol{-}{1,135}
& \toksol{Compute}{1,074}
& \toksol{Me}{10,029}
& \toksol{So}{9,883}
\\

44
& \toksol{Would}{654}
& \toksol{Number}{651}
& \toksol{).}{1,132}
& \toksol{Another}{1,057}
& \toksol{So}{9,689}
& \toksol{Me}{9,846}
\\

45
& \toksol{?}{644}
& \toksol{N}{645}
& \toksol{Total}{1,114}
& \toksol{N}{1,053}
& \toksol{\textasciicircum}{9,680}
& \toksol{\textasciicircum}{9,764}
\\

46
& \toksol{X}{638}
& \toksol{Would}{638}
& \toksol{3}{1,113}
& \toksol{Are}{1,027}
& \toksol{M}{9,002}
& \toksol{M}{9,106}
\\

47
& \toksol{Problem}{631}
& \toksol{Problem}{604}
& \toksol{Not}{1,092}
& \toksol{Not}{1,027}
& \toksol{R}{8,640}
& \toksol{R}{8,738}
\\

48
& \toksol{N}{628}
& \toksol{Total}{599}
& \toksol{Compute}{1,072}
& \toksol{Total}{1,013}
& \toksol{In}{7,900}
& \toksol{In}{7,964}
\\

49
& \toksol{Number}{582}
& \toksol{*}{583}
& \toksol{Check}{1,039}
& \toksol{-}{1,006}
& \toksol{P}{7,598}
& \toksol{T}{7,697}
\\

50
& \toksol{*}{575}
& \toksol{Can}{568}
& \toksol{Another}{1,011}
& \toksol{Check}{987}
& \toksol{T}{7,556}
& \toksol{P}{7,676}
\\

\bottomrule
\end{tabular*}
\end{table*}

\paragraph{Sensitive Token Types under Alternative Prompts.}
Tables~\ref{tab:answer_level_top50_shift_tokens}, \ref{tab:outcome_only_top50_shift_tokens}, and~\ref{tab:reference_solution_top50_shift_tokens} reveal a highly consistent lexical pattern across all three counterfactual prompt constructions. Under both positive and negative privileged conditions, the large-shift lists are dominated by sentence starters, discourse markers, function words, and other locally substitutable surface-form tokens, including \textit{The}, \textit{But}, \textit{So}, \textit{Therefore}, \textit{Wait}, \textit{Alternatively}, \textit{Let}, \textit{Maybe}, and \textit{Check}. Many of the same token identities also occur frequently in both the positive- and negative-shift tails, showing that shift sign does not isolate distinct classes of outcome-supporting and outcome-opposing tokens. This pattern remains evident under reference-solution conditioning, even though the positive and negative privileged contexts contain complete reasoning processes with opposing correctness polarity.

By contrast, the near-zero lists are dominated by digits, arithmetic operators, mathematical delimiters, variables, and other tokens that directly participate in problem-specific expressions, including \(0\)--\(9\), \(+\), \(-\), \(=\), \(/\), parentheses, square-root and exponent symbols, and variables such as \(X\), \(Y\), \(N\), and \(K\). Some punctuation tokens appear in both regimes because their functional roles depend on context: periods and commas may serve as sentence-level punctuation in high-shift positions but as decimal points or digit separators in low-shift mathematical expressions. Overall, the qualitative separation observed under the polarized prompts persists across answer-level, mild outcome-only, and reference-solution conditioning. Large privileged shifts therefore more consistently identify sensitivity to counterfactual framing and stylistic realization than greater problem-specific reasoning value.

\FloatBarrier
\setcounter{section}{4}

\section{Representative Failure Cases of Shift-Directed GRPO}
\label{app:sd_grpo_failure_cases}

The three SD-GRPO variants with $\lambda\in\{0.05,0.1,0.2\}$ exhibit the same qualitative failure mode, despite modifying different fractions of token-level advantages. Rather than separating the cases by threshold, we present two representative examples from AIME24 in Table~\ref{tab:sd_grpo_failure_cases}. The response excerpts are shortened for readability, and only the first two occurrences of each repeated span are retained.

\begin{table*}[t]
\centering
\caption{
Representative degenerate generations produced by SD-GRPO on AIME24. In both examples, initially plausible reasoning transitions into an indefinitely repeated span beginning with \textit{The}.
}
\label{tab:sd_grpo_failure_cases}
\scriptsize
\setlength{\tabcolsep}{4.0pt}
\renewcommand{\arraystretch}{1.12}

\begin{tabularx}{\textwidth}{
@{}
>{\centering\arraybackslash}p{0.055\textwidth}
>{\raggedright\arraybackslash}p{0.32\textwidth}
>{\raggedright\arraybackslash}X
@{}
}
\toprule
\textbf{Case}
& \textbf{Question}
& \textbf{Response Excerpt and Repetition Onset}
\\
\midrule

1
&
Among the 900 residents of Aimeville, 195 own a diamond ring, 367 own a set of golf clubs, and 562 own a garden spade. Each resident also owns a bag of candy hearts. There are 437 residents who own exactly two of these things and 234 who own exactly three. Find the number who own all four.
&
Okay, let's start with the numbers. The total number of residents is 900. The problem states that 195 own a diamond ring, 367 own golf clubs, and 562 own a garden spade. Each resident also owns a bag of candy hearts. So, the first step is to calculate the number of residents who own exactly two items. The 437 residents are $437/900=0.4855\ldots$ of the total, and the 234 residents are $234/900=0.26\ldots$ of the total.

\medskip
\textbf{Repetition onset:}
\texttt{The 0.4855... is the percentage of residents who own exactly two items, and the 0.26... is the percentage who own exactly three items.}

\texttt{The 0.4855... is the percentage of residents who own exactly two items, and the 0.26... is the percentage who own exactly three items.}

\textit{[The same span continues repeatedly.]}
\\

\addlinespace[4pt]

2
&
Jen selects four distinct numbers from
$S=\{1,2,\ldots,10\}$, after which four numbers are chosen uniformly at random. She wins a prize if at least two of her selected numbers are chosen and the grand prize if all four are chosen. If the conditional probability of winning the grand prize given that she won a prize is $m/n$, find $m+n$.
&
Okay, let's break down the problem step by step. The key is to find the probability of Jen winning the grand prize given that she won a prize. The first step is to calculate the probability of her winning the grand prize, which is $1/10$. Then, we need to compare this with the probability of her winning a prize, which is $1/10$. The next step is to compare the two probabilities: $1/10$ versus $1/10$.

\medskip
\textbf{Repetition onset:}
\texttt{The final answer is 1/10, which is 10/100.}

\texttt{The final answer is 1/10, which is 10/100.}

\textit{[The same span continues repeatedly.]}
\\

\bottomrule
\end{tabularx}
\end{table*}

Inspection of all 30 AIME24 validation responses from the collapsed SD-GRPO runs reveals the same characteristic pattern: the generation eventually enters a repetitive loop whose recurring span begins with \textit{The}. This behavior is consistent with the privileged-shift statistics reported in the main text. Under the positive counterfactual condition, \textit{The} appears $8{,}328$ times among sampled-token occurrences satisfying $z^{\mathrm{pos}}>0.1$, compared with $4{,}652$ occurrences satisfying $z^{\mathrm{pos}}<-0.1$. The answer-conditioned privileged prompt used by SD-GRPO exhibits the same qualitative positive-shift imbalance.

Because SD-GRPO replaces the verifier-derived advantage at selected positions with
$\operatorname{sign}(z_{i,t})|A_i|$, occurrences of \textit{The} are more frequently assigned positive than negative token-level advantages. Repeated reinforcement of this common surface-form token increases the policy's tendency to generate it, after which autoregressive feedback amplifies the pattern into persistent repetition. The fact that the same collapse occurs for all three thresholds shows that the failure is not specific to selecting too many tokens. Instead, even a small subset of privileged shift directions can systematically reinforce high-frequency surface-form tokens and destabilize generation.

\end{document}